\documentclass[journal]{IEEEtran}

\usepackage{cite}
\ifCLASSINFOpdf
  \usepackage[pdftex]{graphicx}
\else
\fi
\usepackage{amsmath}
\usepackage{amssymb}
\usepackage{amsthm}

\newcommand{\bomega}{\mathbf{\Omega}}
\newcommand{\ncal}{\mathcal{N}}
\newcommand{\ucal}{\mathcal{U}}

\usepackage[lined,boxed,commentsnumbered,ruled]{algorithm2e} %for algorithm
\usepackage{algorithmic}
\usepackage{array}
\newcolumntype{P}[1]{>{\centering\arraybackslash}p{#1}}
\newcolumntype{M}[1]{>{\centering\arraybackslash}m{#1}}
\usepackage{multirow}
\usepackage{url}
\usepackage[table]{xcolor}
\usepackage{subcaption}

\begin{document}
\title{Few-shots Parallel Algorithm Portfolio Construction via Co-evolution}

\author{Ke~Tang,~\IEEEmembership{Senior Member,~IEEE,}
        Shengcai~Liu,~\IEEEmembership{Member,~IEEE,}
        Peng~Yang,~\IEEEmembership{Member,~IEEE,}
        and~Xin~Yao,~\IEEEmembership{Fellow,~IEEE}% <-this % stops a space
\thanks{The authors are with the Guangdong Key Laboratory of Brain-Inspired Intelligent Computation, Department of Computer Science and Engineering, Southern University of Science and Technology, Shenzhen 518055, China (e-mail: \{tangk3,liusc3,yangp,xiny\}@sustech.edu.cn).
Ke Tang and Shengcai Liu are also with the Research Institute of Trustworthy Autonomous Systems, Southern University of Science and Technology, Shenzhen 518055, China.
Corresponding author: Ke Tang.}}% <-this % stops a space
\maketitle

\begin{abstract}
Generalization, i.e., the ability of solving problem instances that are not available during the system design and development phase, is a critical goal for intelligent systems.
A typical way to achieve good generalization is to learn a model from vast data.   
In the context of heuristic search, such a paradigm could be implemented as configuring the parameters of a parallel algorithm portfolio (PAP) based on a set of ``training'' problem instances, which is often referred to as PAP construction.
However, compared to traditional machine learning, PAP construction often suffers from the lack of training instances, and the obtained PAPs may fail to generalize well.
This paper proposes a novel competitive co-evolution scheme, named Co-Evolution of Parameterized Search (CEPS), as a remedy to this challenge. 
By co-evolving a configuration population and an instance population, CEPS is capable of obtaining generalizable PAPs with few training instances.
The advantage of CEPS in improving generalization is analytically shown in this paper.
Two concrete algorithms, namely CEPS-TSP and CEPS-VRPSPDTW, are presented for the Traveling Salesman Problem (TSP) and the Vehicle Routing Problem with Simultaneous Pickup–Delivery and Time Windows (VRPSPDTW), respectively.
Experimental results show that CEPS has led to better generalization, and even managed to find new best-known solutions for some instances.
\end{abstract}


% \fi
\IEEEpeerreviewmaketitle

\section{Introduction}
\IEEEPARstart{I}{n} the past decades, search methods have become major approaches for tackling various computationally hard problems.
Most, if not all, established search methods, from specialized heuristic algorithms tailored for a particular problem class, e.g., the Lin–Kernighan (LK) heuristic for the Traveling Salesman Problem (TSP) \cite{LinK73}, to general algorithmic frameworks, e.g., Evolutionary Algorithms, share a common feature.
That is, they are parameterized algorithms, which means they involve parameters that need to be configured by users before the algorithm is applied to a problem. 

Although theoretical analyses for many parameterized algorithms have offered worst or average bounds on their performance, their actual performance in practice is in many cases highly sensitive to the settings of parameters \cite{EibenS11,hutter2011sequential,KarafotiasHE15,HuangLY20}.
More importantly, finding the optimal configuration, i.e., parameter setting, requires knowledge of both the algorithm and the problem to solve, which cannot be done manually with ease.
Hence, a lot of efforts have been made to automate this procedure, often dubbed automatic parameter tuning \cite{DymondEKH15,HuangLY20} when the algorithms have relatively few parameters with mostly real-valued domains, or automatic algorithm configuration \cite{hutter2009paramils,hutter2011sequential,ansotegui2009gender,AnsoteguiMSST15,lopez2016irace} when the algorithms have more types (e.g., ordinal and categorical) of parameters.
These methods essentially involve a high-level iterative generate-and-test process.
To be specific, given a set of instances from the target problem class, different configurations are iteratively generated and tested on the instance set.
Upon termination, the process outputs the configuration that performs the best on the instance set.
Since a configuration fully instantiates a parameterized algorithm, for brevity, henceforth we will use the term ``configuration'' to directly denote the resultant solver specified by it.

Built upon automatic algorithm configuration, the automatic construction of parallel algorithm portfolios (PAPs) \cite{GomesS01,huberman1997economics,LiuT019,PengTCY10,LiuT020} seeks to identify a set of configurations to form a PAP.
Each configuration in the PAP is called a component solver.
To solve a problem instance, all the component solvers are run independently, typically in parallel, to get multiple solutions.
Then, the best solution will be taken as the output of the PAP.
Although a PAP would consume much more computational resources than a single-configuration solver, it has two important advantages.
First, the performance of a PAP on any given instance is the best performance achieved among its component solvers on the instance.
In other words, by exploiting the complementarity between the component solvers, a PAP could achieve a much stronger overall performance than any of its component solver.
Second, considering the great development of parallel computing architectures \cite{asanovic2006landscape} (e.g., multi-core CPUs) over the last decade, exploiting parallelism has become very important in designing efficient solvers for computationally hard problems.
PAPs employ parallel solution strategies, and thus allow exploiting modern high-performance computing facilities in an extremely simple way.
%This merit is sometimes even more important than solution quality, since the wall-clock runtime is always a crucial performance indicator for real-world optimization systems. 

From the practical point of view, a PAP construction method is expected to identify a PAP that generalizes well, i.e., performs well not only on the instance set used during the tuning phase, but also on unseen instances of the same problem class.
The reason is that intelligent systems, which incorporate parameterized search algorithms as a module, are seldom built to address a few specific problem instances, but for a whole target problem class, and it is unlikely to know in advance the exact problem instances that a system will encounter in practice.
The need for generalization requires the instance set used for PAP construction to be sufficiently large such that it consists of good representatives of all instances of the target problem class.
Unfortunately, in a real-world scenario PAP construction is very likely to face the few-shots challenge.
That is, the available instance set is not only of small size, but also may not well represent the target problem class.
For example, the widely studied TSP benchmark suites (i.e., TSPlib \cite{Reinelt91}) consist of a few hundred TSP instances, while there could be millions of possibilities for concrete TSP instances even if only considering a fixed number of cities.
In consequence, the more powerful of a PAP construction method, the higher risk that the obtained PAP will over-fit the instances involved in the tuning process.

This paper suggests that the pursuit of generalizable PAPs could be modeled as a co-evolutionary system, in which two internal populations, representing the configurations (the PAP) and the problem instances, respectively, compete with each other during the evolution course.
The evolution of the latter promotes exploration in the instance space of the target problem class to generate synthetic instances that exploit the weakness of the former.
The former, on the other hand, improves itself by identifying configurations that could better handle the latter.
In this way, the configuration population (the PAP) is encouraged to evolve towards achieving good performance on as many instances of the target problem class as possible, i.e., towards better generalization.
Specifically, contributions of this paper include: 
\begin{enumerate}
\item A novel PAP construction framework, namely Co-Evolution of Parameterized Search (CEPS), is proposed. It is also shown that CEPS approximates a process that minimizes the upper bound, i.e., a tractable surrogate, of the generalization performance.
\item To demonstrate the implementation details of CEPS as well as to assess its potential, concrete instantiations are also presented for two hard optimization problems, i.e., TSP and the Vehicle Routing Problem with Simultaneous Pickup–Delivery and Time Windows (VRPSPDTW) \cite{WangC12}.
Computational studies confirm that CEPS is able to obtain PAPs with better generalization performance.
\item The proposal of CEPS extends the realm of Co-Evolution, for the first time, to evolving algorithm configurations and problem instances. Since CEPS does not invoke domain-specific knowledge, its potential applications can go beyond optimization problems, even to planning and learning problems.
\end{enumerate}

The rest of the paper is organized as follows.
Section~\ref{sec:insight} introduces the challenge of seeking generalizable PAPs, existing PAP construction methods, as well as the theoretical insight behind CEPS.
Section~\ref{sec:ceps} presents the CEPS framework.
Section~\ref{sec:instantiation} presents its instantiations for TSP and VRPSPDTW.
Computational studies on these two problems are presented in Section~\ref{sec:experiment}.
Threats to validity of this study are discussed in Section~\ref{sec:threats}.
Section~\ref{sec:conclusion} concludes the paper with discussions.

\section{Parameterized Solvers Made Generalizable}
\label{sec:insight}
\subsection{Notations and Problem Description}
Assume a PAP is to be built for a problem class (e.g., TSP), for which an instance of the problem class is denoted as $s$, and the set of all possible $s$ is denoted as $\bomega$.
Given a parameterized algorithm, each component solver of the PAP is a configuration (full instantiation) of the algorithm.
Generally speaking, the parameterized algorithm can be any concrete computational process, e.g., a traditional heuristic search process such as the LK Heuristic for TSP or even a neural network \cite{NazariOST18,ChenT19,KoolHW19} that outputs a solution for a given instance of the target problem class.
Let $\theta$ denote a configuration and let $\Theta$ denote a PAP that contains $K$ different configurations (component solvers), i.e., $\Theta=\{\theta_1,...,\theta_K\}$.
The quality of a configuration $\theta$ on a given instance $s$ is denoted as $f(s, \theta)$, which is a performance indicator of the corresponding solver on the instance.
This indicator could concern many aspects, e.g., the quality of the obtained solution \cite{lopez2016irace}, the CPU time required to achieve a solution above a given quality threshold \cite{hutter2009paramils}, or even be stated in a multi-objective form \cite{BlotHJKT16}.
The performance of a PAP $\Theta$ on an instance s, denoted as $f(s, \Theta)$, is the best performance achieved among its component solvers $\theta_1,...,\theta_k$ on $s$ (assuming the smaller $f(s, \theta)$, the better):
\begin{equation}
\label{eq:fsTheta}
f(s, \Theta) := \min \left\{ f(s,\theta_1),...,f(s,\theta_K) \right\}.
\end{equation}

Following the above definitions, optimizing the generalization performance of a PAP can be stated as:
\begin{equation}
\label{eq:optimizegeneralization_1}
  \min _{\Theta} J(\Theta) := \int_{s \in \bomega} f(s, \Theta) p(s) ds,
\end{equation}
where $p(s)$ stands for the prior distribution of $s$.
Since in practice the prior distribution is usually unknown, a uniform distribution can be assumed without loss of generality.
Eq.~(\ref{eq:optimizegeneralization_1}) can be then simplified to Eq.~(\ref{eq:optimizegeneralization_2}) by omitting a normalization constant:
\begin{equation}
\label{eq:optimizegeneralization_2}
  \min _{\Theta} J(\Theta) := \int_{s \in \bomega} f(s, \Theta) ds.
\end{equation}

The challenge with Eqs.~(\ref{eq:optimizegeneralization_1}) and (\ref{eq:optimizegeneralization_2}) is that in practice they cannot be directly optimized since the set $\bomega$ is generally unavailable.
Instead, only a set of ``training'' instances, i.e., a subset $T \subset \bomega$, is given for the purpose of constructing $\Theta$.
In fact, the so-called over-tuning phenomenon \cite{Birattari2004,LiuTL020}, which is analogous to the over-fitting phenomenon in machine learning, has been observed when the size of the training instance set is rather limited (i.e., few-shots challenge).
That is, the test (generalization) performance of the obtained configurations is arbitrarily bad even if their performance on the training set is excellent.
Even worse, given a $T$ collected from real world, it is non-trivial to know how to verify whether it is a good representative of $\bomega$.
In case the training instance set is too small, or is not a good representative of the whole problem class, the best PAP obtained with it would fail to generalize.

\subsection{Related Work}
\label{sec:related_work}
Currently, there exist several approaches for PAP construction, namely GLOBAL \cite{lindauer2017automatic}, PARHYDRA \cite{xu2010hydra,lindauer2017automatic}, CLUSTERING \cite{kadioglu2010isac} and PCIT \cite{LiuT019}.
GLOBAL considers PAP construction as an algorithm configuration problem by treating $\Theta$ as a parameterized algorithm.
By this means existing automatic algorithm configuration tools could be directly utilized to configure all the component solvers of $\Theta$ simultaneously.
In comparison, PARHYDRA constructs $\Theta$ iteratively by identifying a single component solver in each iteration that maximizes marginal performance contribution to the current PAP.
CLUSTERING and PCIT are two approaches based on instance grouping.
That is, they both first split the training set into disjoint subsets, and then identify a component solver on each subset.
The former splits the training set by clustering training instances based on their feature-vector representations, while the latter splits the training set uniform randomly and will adjust the instance grouping by transferring instances between subsets during the PAP construction process.

Other than PAP, another important way of utilizing an algorithm portfolio to achieve stronger overall performance is algorithm selection (AS), which seeks to select, from a given algorithm portfolio, the best suited solver to solve an instance.
Specifically, the algorithm selector is usually built by training machine learning models based on the feature sets of training instances.
Over the last decades, AS has been successfully applied to many computationally hard problems, such as Boolean satisfiability problems \cite{xu2008satzilla} and TSP \cite{KerschkeKBHT18,zhao2020leveraging}. A comprehensive survey on AS can be found in \cite{Kotthoff14}.

To the best of our knowledge, the few-shots challenge has not been investigated yet in the literature.
That is, all the methods mentioned above assume that the training instances could sufficiently represent that target problem class.
However, as aforementioned, such an assumption could not be always true since in some cases, we might only have scarce or biased training instances.

\subsection{Enhancing Generalization with Synthetic Instances}
\label{sec:theoretical_insights}
A natural idea to tackle the few-shots challenge is to augment $T$ with a set of synthetic instances, say $T'$, such that the PAP obtained with $T \cup T'$ would generalize better than that obtained with $T$.
This idea is generally valid because if the size of $T'$ continues to grow, $T \cup T'$ will eventually approach $\bomega$.
Hence, the key question is how a generalizable PAP could be obtained with a sufficiently small $T'$.
This question can be re-stated as: how to generate synthetic training instances, such that the generalization of the obtained PAP could be improved as much as possible with a $T'$ of (say predefined) small size.

Given a parameterized algorithm, suppose a PAP $\Theta$ has been obtained as the best-performing PAP on $T$.
A synthetic instance set $T'$ is to be generated, with the aim that a new PAP $\Theta'$ obtained with $T \cup T'$ would outperform $\Theta$ in terms of generalization as much as possible.
Ignoring the inner optimization/tuning process with which $\Theta'$ and $\Theta$ are obtained, generating high-quality $T'$ could be more formally stated as another optimization problem as in Eq.~(\ref{eq:optimization_formulation}):

\small
\begin{equation}
\label{eq:optimization_formulation}
\begin{split}
  &\min_{T'} \left\{J(\Theta')-J(\Theta)\right\}:=\int_{s \in \bomega} f\left(s, \Theta'\right) d s-\int_{s \in \bomega} f \left(s, \Theta \right) d s\\
  &=\left[ \sum_{s \in T} f(s, \Theta')+ \sum_{s \in T'} f(s, \Theta')+\int_{s \in \bomega \backslash (T \cup T')} f(s, \Theta') d s\right]\\
  &-\left[\sum_{s \in T} f(s, \Theta) + \sum_{s \in T'} f(s, \Theta) + \int_{s \in \bomega \backslash (T \cup T')} f(s, \Theta) d s\right].
\end{split}
\end{equation}
\normalsize
Eq.~(\ref{eq:optimization_formulation}) aims at achieving the largest improvement over $J(\Theta)$, which is a constant since $\Theta$ has been obtained with $T$.
Since $\Theta'$ is obtained with $T \cup T'$, we further assume that for any $s \in \bomega$, $f(s, \Theta')\leq f(s, \Theta)$.
Although this is a rather restrictive assumption, it will be shown later that it could be fulfilled when $\Theta$ is a subset of $\Theta'$.
Applying this assumption to the right hand side of Eq.~(\ref{eq:optimization_formulation}), we have:
\begin{equation}
\label{eq:break_down}
\begin{split}
& \sum_{s \in T} [ f \left(s, \Theta' \right) - f(s, \Theta) ] \leq 0, \\
& \sum_{s \in T'} [ f \left(s, \Theta' \right) - f(s, \Theta) ] \leq 0, \\
& \int_{s \in \bomega \backslash (T \cup T')} [ f \left(s, \Theta' \right) - f(s,\Theta) ] d s \leq 0.
\end{split}
\end{equation}
Considering that in the right hand side of Eq.~(\ref{eq:optimization_formulation}), $\bomega \backslash (T \cup T')$ is unknown, we thus discard the terms regarding $\bomega \backslash (T \cup T')$ and retain the ones regarding $T$ and $T'$.
By inequality~(\ref{eq:break_down}), the discarded terms are non-positive, we then have:
\begin{equation}
\label{eq:upper_bound}
 J \left(\Theta'\right)-J(\Theta) \leq \sum_{s \in T \cup T'} \left[ f\left(s, \Theta'\right)-f(s, \Theta) \right].
\end{equation}

Inequality~(\ref{eq:upper_bound}) gives an upper bound of $J\left(\Theta'\right)-J(\Theta)$ that depends on the current instance set $T$, the target instance set $T'$, the current PAP $\Theta$, and the new PAP $\Theta'$.
Note the upper bound is always non-positive by Inequality~(\ref{eq:break_down}), which means if the assumption holds, the new PAP $\Theta'$ is guaranteed to generalize better than the current PAP $\Theta$.
More importantly, considering that neither $J\left(\Theta'\right)$ nor $J(\Theta)$ can be precisely measured in practice, the upper bound in Inequality~(\ref{eq:upper_bound}) provides a measurable surrogate for minimizing $J(\Theta')-J(\Theta)$, such that even larger performance improvement could be achieved than only relying on the assumption.
Therefore, given a training instance set $T$ and a PAP $\Theta$ obtained with $T$, an improved PAP $\Theta'$ (in terms of generalization performance) could be obtained with a strategy with two steps to minimize the upper bound in Inequality~(\ref{eq:upper_bound}):
\begin{enumerate}
  \item identify the $T'$ that maximizes $\sum_{s \in T \cup T'} f(s, \Theta)$ (this is equivalent to maximizing $\sum_{s \in T'} f(s, \Theta)$ since $\sum_{s \in T} f(s, \Theta)$ is a constant given that $T$ and $\Theta$ are fixed);
  \item identify the $\Theta'$ that minimizes $\sum_{s \in T \cup T'} f\left(s, \Theta' \right)$  (note that, once $T'$ is generated, the term $\sum_{s \in T \cup T'} f(s, \Theta)$ is a constant and can be omitted).
\end{enumerate}

The above two steps naturally serve as the core building-block of an iterative process that gradually seek PAPs with better generalization performance.
There could be many ways to design such an iterative process.
Among them Competitive Co-evolution \cite{RosinB97} provides a readily available framework.
That is, one can maintain an instance population (representing the instance set) and a configuration population (representing the PAP).
In each iteration, the two populations alternately evolve and compete with each other, i.e., the instance population evolves to identify $T'$ and the configuration population evolves to identify $\Theta'$.

Recall that the two-step improvement strategy is derived from the assumption that $f(s, \Theta')\leq f(s, \Theta)$ for any $s \in \bomega$.
This assumption holds if $\Theta$ is a subset of $\Theta'$ because by definition of PAP (Eq.~(\ref{eq:fsTheta})), we have:
$f(s,\Theta')=\min \{ f(s,\Theta), \min_{\theta \in \Theta' \backslash \Theta}f(s, \theta)\} \leq f(s, \Theta)$.
Following this, one could further design the evolution of the PAP (the configuration population) as identifying new configurations to insert into the current PAP $\Theta$, such that the new PAP $\Theta'$, which is a superset of $\Theta$, minimizes $\sum_{s \in T \cup T'} f\left(s, \Theta' \right)$.
However, in practice such a mechanism could suffer from the PAP-size issue.
That is, the number of the component solvers in the PAP will keep increasing as the co-evolution proceeds.
Recall that a PAP runs its component solvers in parallel; thus its size is mandatorily limited by the available computational resources (e.g., the number of available CPU cores) and thus cannot grow infinitely.
A natural way to avoid this issue is to first remove some configurations from the PAP $\Theta$, resulting in a temporary PAP $\bar{\Theta}$, and then identify new configurations to insert into $\bar{\Theta}$, such that the final PAP $\Theta'$ is of the same size as $\Theta$.
However, this approach no longer guarantees the validity of the above assumption.
As a consequence, $\Theta'$ may generalize worse than $\Theta$.
A remedy to prevent this as much as possible is to increase redundancy in the evolution of the PAP.
More specifically, one could repeat the configuration-removal procedure to $\Theta$ for $n$ times, leading to $n$ temporary PAPs, $\bar{\Theta}_1,...,\bar{\Theta}_n$;
then for each temporary PAP $\bar{\Theta}$, the new configurations are identified and inserted, leading to $n$ new PAPs, $\Theta'_1,...,\Theta'_n$, each of which is of the same size as $\Theta$;
finally, the PAP among them that performs best against $T \cup T'$ is retained.

\section{Co-evolution of Parameterized Search}
\label{sec:ceps}
By incorporating the above-described procedure into the co-evolution process, we arrive at the proposed CEPS framework, as demonstrated in Algorithm~\ref{alg:ceps}.
In general, CEPS consists of two major phases, i.e., an initialization phase (lines 2-7), and a co-evolution phase (lines 8-27) which could be further subdivided into 
alternating between the evolution of the configuration population (representing the PAP) (lines 10-15) and the evolution of the instance population (representing the training instances) (lines 17-26) for \textit{MaxIte} iterations in total.
These modules are detailed as follows.

\begin{algorithm}[tbp]
  \LinesNumbered
  \SetKwInOut{Input}{input}
  \SetKwInOut{Output}{output}
  \Input{training set $T$; number of component solvers, $K$; number of temporary PAPs, $n$; maximum number of iterations, $MaxIte$}
  \Output{the final configuration population (PAP) $\Theta$}
  \tcc{--------Initialization--------}
  Randomly sample a set $C$ from the configuration space, and test all the selected configurations on $T$;\\
  $\Theta \leftarrow \varnothing$;\\  
  \For{$i \leftarrow 1$ \KwTo $K$}
  {
    Find $\theta_i$ from $C$, with the target minimizing $\frac{1}{|T|} \Sigma_{s\in T} f\left(s, \Theta \cup\{ \theta_i\}\right)$;\\
    $\Theta \leftarrow \Theta \cup \{\theta_i\}$;\\
  }
  \For{$ite \leftarrow 1$ \KwTo $MaxIte$}
  {
    \tcc{-------Evolution of $\Theta$-------}
    \For{$i \leftarrow 1$ \KwTo $n$}
    {
      Randomly select $\theta \in \Theta$, and let $\bar{\Theta}_i \leftarrow \Theta \backslash \{\theta\}$;\\
      Use SMAC to identify $\theta'$, with the target minimizing $\frac{1}{|T|} \Sigma_{s\in T} f\left(s, \bar{\Theta}_i \cup\{ \theta'\}\right)$;\\
      $\Theta'_i \leftarrow \bar{\Theta}_i \cup\{ \theta'\}$;\\
    }
    $\Theta \leftarrow$ the best-performing PAP among $\Theta'_1,...,\Theta'_n$;\\
    \tcc{-------Evolution of $T$-------}
    \lIf{$ite=MaxIte$}{break}
    $T' \leftarrow $ create a copy of $T$;\\
    Assign the fitness of each $s \in T'$ as $f(s, \Theta)$;\\
    \While{not terminated}
    {
      $s'\leftarrow$ randomly select $s \in T'$, and \textbf{mutate} $s$;\\
      Test $s'$ with $\Theta$ and assign the fitness of $s'$ as $f(s', \Theta)$;\\
      $s^* \leftarrow $ randomly select one from all the instances in $T'$ with lower fitness than $s'$;\\
      $T \leftarrow T \backslash \{s^*\} \cup \{s'\}$;\\
    }
    $T \leftarrow T' \cup T$;\\
  }
\Return{$\Theta$}
\caption{The General Framework of CEPS}
\label{alg:ceps}
\end{algorithm}

\subsubsection{Initialization}
Given an initial training instance set $T$, a simple greedy strategy is adopted to initialize a configuration population (PAP) $\Theta$ of size $K$.
First, a set of candidate configurations $C$ are randomly sampled from the configuration space and tested on the training set $T$ (line 2).
Then, starting from an empty set (line 3), $\Theta$ is built iteratively (lines 4-7).
At each iteration, the configuration whose inclusion into $\Theta$ leads to the largest performance improvement is selected from $C$ (line 5) and inserted into $\Theta$ (line 6).
The process terminates when $K$ configurations have been selected.

\subsubsection{Evolution of the Configuration Population}
Given a configuration population $\Theta$, $n$ temporary PAPs, $\bar{\Theta}_1,...,\bar{\Theta}_n$, are first generated by repeatedly randomly removing a configuration from $\Theta$ (line 11).
Then for each $\bar{\Theta}_i$, an existing automatic algorithm configuration tool, namely SMAC \cite{hutter2011sequential}, is used to search in the configuration space to find a new configuration $\theta'$ with the target that the inclusion of $\theta'$ into $\bar{\Theta}_i$ leads to the minimization of the performance of the resultant PAP $\Theta'_i$ on the training set (line 12-13). 
Finally, the best-performing PAP among the $n$ new PAPs $\Theta'_1,...,\Theta'_n$ will replace $\Theta$ (line 15).
From the perspective of evolutionary computation, the above procedure could be seemed as mutation to $\Theta$, with SMAC employed as the mutation operator.

\subsubsection{Evolution of the Instance Population}
In this phase CEPS first creates a copy of $T$, i.e., $T'$, that will serve as the initial instance population hereafter (line 18).
Since the aim of the evolution of the instance population is to identify a $T'$ that are hard for $\Theta$, i.e., maximizing $\sum_{s \in T'} f(s, \Theta)$,
each instance in $T'$ is assigned with a fitness as the performance of $\Theta$ on it (line 19) — the worse the performance, the higher the fitness.
In each generation of the evolution of the instance population, CEPS first randomly selects an instance $s$ from $T'$ as the parent and mutates it to generate an offspring $s'$ (line 21), which is then evaluated against the configuration population (line 22).
Finally, CEPS uses $s'$ to randomly replace an instance in $T'$ that has lower fitness than $s'$ (lines 23-24).
In this way, as the number of generations increases, the average fitness of instances in $T'$ will gradually increase, meaning that the instances in $T'$ will be harder and harder for the configuration population.
When the evolution of the instance population ends, the final $T'$ will be merged into the training set (line 26), which will be used for obtaining $\Theta'$ in the next iteration of the co-evolution.
Note in the last iteration (i.e., the \textit{MaxIte}-th iteration) of the co-evolution phase, evolution of the instance population is skipped (line 17) because there is no need to generate more instances since the final configuration population has been constructed completely.

\begin{algorithm}[tbp]
  \LinesNumbered
  \SetKwInOut{Input}{input}
  \SetKwInOut{Output}{output}
  \Input{instance $s$}
  \Output{mutated instance $s$}
  Let $N$ be the number of cities in $s$, which is then represented by $\{(x_1,y_1),...,(x_N,y_N)\}$;\\
  $x_{min} \leftarrow \min \{x_1,...,x_N\}$; $x_{max} \leftarrow \max \{x_1,...,x_N\}$;\\
  $y_{min} \leftarrow \min \{y_1,...,y_N\}$; $y_{max} \leftarrow \max \{y_1,...,y_N\}$;\\
  \For{$i \leftarrow 1$ \KwTo $N$}
  {
    Generate a random number $r \in[0, 1]$;\\
    \uIf{$r \leq 0.9$}
    {
    Sample $\Delta \sim \ncal\left(0,[0.025\cdot(x_{max}-x_{min})]^2\right)$;\\
    $x_i \leftarrow x_i + \Delta$;\\
    Sample $\Delta \sim \ncal\left(0,[0.025\cdot(y_{max}-y_{min})]^2\right)$;\\
    $y_i \leftarrow y_i + \Delta;$\\
    }
    \Else
    {
    Sample $x' \sim \ucal(x_{min}, x_{max})$;\\
    $x_{i} \leftarrow x';$\\
    Sample $y' \sim \ucal(y_{min}, y_{max})$;\\
    $y_{i} \leftarrow y';$\\
    }
  }
\Return{$s$}
\caption{The Instance Mutation Operator in CEPS-TSP}
\label{alg:mutator-tsp}
\end{algorithm}

\section{Instantiations for TSP and VRPSPDTW}
\label{sec:instantiation}
Algorithm~\ref{alg:ceps} is a rather generic framework since the representations of both populations depend on the target parameterized algorithm and the target problem class, respectively.
The mutation operator for the instance population, as well as the fitness function also depend on target problem class.
In this paper, two instantiations of CEPS, namely CEPS-TSP and CEPS-VRPSPDTW, have been developed for the TSP and VRPSPDTW problems, respectively.
These two target problem classes are chosen because, as a classic NP-hard problem, TSP is one of the most widely investigated benchmarking problems in academia.
In comparison, VRPSPDTW is a much more complex routing problem that takes real-world requirements into account.
The significant difference between these two problems could provide a good context for assessing CEPS.

\subsection{CEPS-TSP}
\label{sec:ceps_tsp}

Given a list of cities and the distances between each pair of cities, the target of TSP is to find the shortest route that visits each city and returns to the origin city. 
Specifically, the symmetric TSP with distances in a two-dimensional Euclidean space is considered here.
\subsubsection{Instance Mutation Operator}
Each of such TSP instance is represented by a list of $(x,y)$ coordinates with each coordinate as a city.
An operator widely used for generating TSP instances (see \cite{Hemert06}), is employed as the instance mutation operator of CEPS-TSP.
As illustrated in Algorithm~\ref{alg:mutator-tsp}, the mutation operator works as follows.
Let $x_{min}$ and $x_{max}$, $y_{min}$ and $y_{max}$, be the minimum and the maximum of the ``$x$'' values and the ``$y$'' values across all coordinates of a given instance $s$, respectively.
When applying mutation to $s$, for each coordinate $(x,y)$ in $s$, $x$ and $y$ are offset with probability 0.9 by the step sizes sampled from $\ncal\left(0,[0.025\cdot(x_{max}-x_{min})]^2\right)$ and $\ncal\left(0,[0.025\cdot(y_{max}-y_{min})]^2\right)$, respectively, and with probability 0.1, $x$ and $y$ are replaced by new values sampled from $\ucal(x_{min},x_{max})$ and $\ucal(y_{min},y_{max})$, respectively. $\ncal(\mu, \sigma^2)$ refers to normal distribution with mean $\mu$  and variance $\sigma ^{2}$,
and $\ucal(a,b)$ refers to normal distribution defined on closed interval $[a,b]$.
\subsubsection{Parameterized Algorithm}
The adopted parameterized algorithm is the Helsgaun’s Lin-Kernighan Heuristic (LKH) \cite{Helsgaun09} version 2.0.7 (with 23 parameters), one of the state-of-the-art inexact solver for such TSP.
\subsubsection{Fitness Function}
For TSP, the penalized average runtime with penalty factor 10 (PAR-10) \cite{hutter2009paramils}, is considered as the performance indicator.
The smaller the PAR-10, the better.
More specifically, the performance of a configuration $\theta$ on an instance $s$, i.e., $f(s, \theta)$, is the penalized runtime needed by $\theta$ to solve $s$.
In particular, when running $\theta$ on $s$, the run would be terminated as soon as the optimal solution of $s$ is found or after a cutoff time of 10 seconds.
In the first case, the run is considered successful and $f(s, \theta)$ is exactly the recorded runtime; in the second case, the run is considered timeout and $f(s, \theta)$ is the cutoff time multiplied by the penalty factor 10, i.e., 10 seconds $\times$ 10 = 100 seconds.
Based on $f(s, \theta)$, the performance of a PAP solver $\Theta$ on an instance $s$, i.e., $f(s, \Theta)$, as defined in Eq.~(\ref{eq:fsTheta}), is $\min\{ f(s, \theta)| \theta \in \Theta\}$, which is the fitness function used in the evolution of the instance population in CEPS-TSP (line 19 and line 22 in Algorithm~\ref{alg:ceps}).
Finally, the performance of a solver (a single configuration or a PAP solver) on an instance set is the average of the penalized runtime over all instances in the set, which is directly used for fitness evaluation in CEPS-TSP to compare PAPs constructed with the configuration population (line 15 of Algorithm~\ref{alg:ceps}).

\subsection{CEPS-VRPSPDTW}
\label{sec:ceps_vrpsdptw}
Given a number of customers who require both pickup service and delivery service within a certain time window, the target of VRPSPDTW \cite{WangC12} is to send out a fleet of capacitated vehicles, which are stationed at a depot, to meet the customer demands with the minimum total cost.
More specifically, VRPSPDTW is defined on a complete graph $G=(V,E)$ with $V=\{0,1,2,...,N\}$ as the node set and $E$ as the arc set defined between each pair of nodes, i.e., $E=\{\langle i,j \rangle |i,j \in V, i\neq j\}$.
For convenience, the depot is denoted as 0 and the customers are denoted as $1,...,N$.
Each node $i \in V$ has a coordinate $(x_i,y_i)$
and the distance between $i$ and $j$, denoted as $c_{i,j}$, is the Euclidean distance.
In addition to the coordinate, each customer is associated with 5 attributes, i.e., a delivery demand $d_i$, a pickup demand $p_i$, a time window $[a_i, b_i]$ and a service time $s_i$.
$d_i$ represents the amount of goods to deliver from the depot to customer $i$ and $p_i$ represents the amount of goods to pick up from customer $i$ to be delivered to the depot.
$a_i$ and $b_i$ define the start and the end of the time window in which the customer receives service. The time windows are treated as hard constraints.
That is, arrival of a vehicle at the customer $i$ before $a_i$ results in a wait before service can begin; while arrival after $b_i$ is infeasible.
Finally, $s_i$ is the time spent by the vehicle to load/unload goods at customer $i$.
A fleet of $J$ identical vehicles, each with a capacity of $Q$ and dispatching cost $c_d$, is initially located at the depot.
Each vehicle starts at the depot and then serve the customers, and finally returns to the depot.
For convenience, the depot 0 is also associated with 5 attributes, in which $a_0$ and $b_0$ are the earliest time the vehicles can depart from the depot and the latest time the vehicles can return to the depot, respectively, and $d_0=p_0=s_0=0$.

A solution $S$ to VRPSPDTW could be represented by a set of vehicle routes, i.e., $S=\{R_1,R_2,...,R_K\}$, in which each route $R_i$ consists of a sequence of nodes that the vehicle visits, i.e., $R_i=(h_{i,1},h_{i,2},...,h_{i, L_i})$, where $h_{i,j}$ is the $j$-th node visited in $R_i$, and $L_i$ is the length of $R_i$.
Let $TD(R_{i})$ denote the total travel distance in $R_i$, and let $load(R_i)$ denote the highest load on the vehicle that occurs in $R_i$.
Let $arr(h_{i,j})$ and $dep(h_{i,j})$ denote the time of arrival at $h_{i,j}$ and the time of departure from $h_{i,j}$, respectively.

The total cost of $S$ consists of two parts: the dispatching cost of the used vehicles, which is $K \cdot c_d$, and the transportation cost, which is the total travel distance in $S$ multiplied by unit transportation cost $u$.
The objective of the VRPSPDTW problem is to find routes for vehicles that serve all the customers at a minimal cost, as presented in Eq.~(\ref{eq:vrpspdtw}):
\begin{equation}
\label{eq:vrpspdtw}
\begin{split}
\min_{S} &TC(S):= \sum_{i=1}^K \left[ c_d + TD(R_i) \cdot u \right] \\
s.t.:& K \leq J\\
& h_{i,1}=h_{i,L(i)}=0, 1 \leq i \leq K\\
& \sum_{i=1}^K \sum_{j=2}^{L_i-1} I[h_{i,j} = e] = 1, 1 \leq e \leq N\\
& load(R_i) \leq Q, 1 \leq i \leq K \\
& dep(h_{i,1}) \geq a_0, 1 \leq i \leq K \\
& arr(h_{i,j}) \leq b_{h_{i,j}}, 1 \leq i \leq K, 2 \leq j \leq L_i \\
\end{split},
\end{equation}
where the constraints are: 1) the number of used vehicles must be smaller than the number of available ones; 2) each customer must be served exactly once; 3) the vehicle cannot be overloaded during transportation; 4) the vehicles can only serve after the start of the time window of the depot, and must return to the depot before
the end of the time window of the depot; 5) the service
of the vehicle to each customer must be performed within
that customer’s time window.

We consider a practical application scenario from the JD logistics company.
Consider a VRPSPDTW solver that needs to solve a VRPSPDTW instance every day.
The company has about 3000 customers in total in the city, but only about 13\% of its customers (i.e., 400) require service per day.
Therefore, for the solver, the different VRPSPDTW instances it faces have the following connections: 1) the location and the time window of the depot are unchanged, and the vehicle fleet is unchanged; 2) the locations of the customers will change; 3) the time windows of the customers will change; 4) the delivery and pickup demands of the customers will change.

\begin{algorithm}[tbp]
  \LinesNumbered
  \SetKwInOut{Input}{input}
  \SetKwInOut{Output}{output}
  \Input{instance $s$}
  \Output{mutated instance $s$}
  Let $N$ be the number of customers, which are represented by \footnotesize $\{(x_0,y_0,d_0,p_0,a_0,b_0,s_0),...,(x_N,y_N,d_N,p_N,a_N,b_N,s_N)\}$;\\
  \normalsize
  $x_{min} \leftarrow \min \{x_1,...,x_N\}$; $x_{max} \leftarrow \max \{x_1,...,x_N\}$;\\
  $y_{min} \leftarrow \min \{y_1,...,y_N\}$; $y_{max} \leftarrow \max \{y_1,...,y_N\}$;\\
  $p_{min} \leftarrow \min \{p_1,...,p_N\}$; $p_{max} \leftarrow \max \{p_1,...,p_N\}$;\\
  $u_{min} \leftarrow \min \{u_1,...,u_N\}$; $u_{max} \leftarrow \max \{u_1,...,u_N\}$;\\  
  \For{$i \leftarrow 1$ \KwTo $N$}
  {
  	\tcc{-----Coordinate Mutation-----}
    Generate a random number $r \in[0, 1]$;\\
    \uIf{$r \leq 0.9$}
    {
    Sample $\Delta \sim \ncal\left(0,[0.025\cdot(x_{max}-x_{min})]^2\right)$;\\
    $x_i \leftarrow x_i + \Delta$;\\
    Sample $\Delta \sim \ncal\left(0,[0.025\cdot(y_{max}-y_{min})]^2\right)$;\\
    $y_i \leftarrow y_i + \Delta;$\\
    }
    \Else
    {
    Sample $x' \sim \ucal(x_{min}, x_{max})$;\\
    $x_{i} \leftarrow x';$\\
    Sample $y' \sim \ucal(y_{min}, y_{max})$;\\
    $y_{i} \leftarrow y';$\\
    }
    \tcc{-----Demand Mutation---------}
    Sample $p' \sim \ucal(p_{min}, p_{max})$;\\
    $p_{i} \leftarrow p';$\\
    Sample $d' \sim \ucal(d_{min}, d_{max})$;\\
    $d_{i} \leftarrow d';$\\
    \tcc{-----Time-window Mutation----}
    Sample $\Delta_1, \Delta_2 \sim \ncal\left(0,(0.025 \cdot (b_0 - a_0)^2 \right)$;\\
    $a_i \leftarrow a_i + \Delta_1$;\\
    $b_i \leftarrow b_i + \Delta_2$;\\
  }
\Return{$s$}
\caption{The Instance Mutation Operator in CEPS-VRPSPDTW}
\label{alg:mutator-vrpspdtw}
\end{algorithm}

\subsubsection{Instance Mutation Operator}
Based on the above observation, we design a specialized mutation operator for VRPSPDTW, as presented in Algorithm~\ref{alg:mutator-vrpspdtw}.
First, the coordinate mutation used in CEPS-TSP is also used here.
Moreover, for the pickup demand $p_i$ and the delivery demand $d_i$ of each customer, they are replaced by new values sampled from $\ucal(p_{min}, p_{max})$ and $\ucal(d_{min}, d_{max})$, respectively, where $p_{min}$ and $p_{max}$, $u_{min}$ and $u_{max}$, are the minimum and the maximum of the ``$p$'' value and the ``$d$'' values across all customers of $s$, respectively.
For the time window $[a_i, b_i]$ of each customer, $a_i$ and $b_i$ are offset by the step sizes sampled from $\ncal\left(0,(0.025 \cdot (b_0 - a_0)^2 \right)$, where $a_0$ and $b_0$ are the earliest time that the vehicles can depart from the depot and the latest time that the vehicles can return to the depot.

\subsubsection{Parameterized Algorithm}
The adopted parameterized algorithm for VRPSPDTW is a powerful co-evolutionary genetic algorithm (Co-GA) proposed by \cite{WangC12} (with 12 parameters).

\subsubsection{Fitness Function}
For VRPSPDTW, the penalized average normalized cost (PANC), is considered as the performance indicator. 
The smaller the PANC, the better.
More specifically, the performance of a configuration $\theta$ on an instance $s$, i.e., $f(s, \theta)$, is the penalized normalized cost of the solution found by $\theta$.
In particular, the run of $\theta$ on $s$ would be terminated after a cutoff time of 150 seconds.
Assume $\theta$ successfully finds a feasible solution of cost $c$ to $s$.
Considering for different VRPSPDTW instances, the scales of the solution costs may vary significantly, thus the ``normalized cost'' is introduced to replace $c$:
\begin{equation}
  f(s, \theta) = \frac{c}{mean\_distance(s)},
\end{equation}
where $mean\_distance(s)$ is the average distance between all pairs of customers in instance $s$.
In case that $\theta$ fails to find a feasible solution to $s$ within the cutoff time, the corresponding run is considered timeout and $f(s, \theta)$ will be set to a large penalty value, i.e., 2000.
Based on $f(s, \theta)$, the further definitions of the performance of a PAP solver on an instance (used as the fitness function in the evolution of the instance population), and the performance of a solver on an instance set (used for fitness evaluation to compare different PAPs), are analogous to the case of TSP.

\begin{table*}[hbt!]
  \centering
  \caption{Summary of the experimental settings.}
  \scalebox{0.92}{
  \begin{tabular}{M{.2\columnwidth}M{.39\columnwidth}M{.21\columnwidth}M{.65\columnwidth}M{.35\columnwidth}}
  \hline
            & Instance Sets & \#solvers in PAP & Performance Indicator & Parameterized Algorithm\\
  \hline
  TSP & 500 instances generated by 10 different generators. Training/Testing split: 30/470 & 4 & Runtime needed to find the optima of the instances. In particular, PAR-10 with cutoff time 10 seconds was used (see Section~\ref{sec:ceps_tsp}) & LKH version 2.0.7 \cite{Helsgaun09} with 23 parameters \\
  \hline
  VRPSPDTW & 233 instances obtained from real-world application. Training/Testing split: 14/219 & 4 & Cost of the found solutions. In particular, PANC with cutofftime 150 seconds was used (see Section~\ref{sec:ceps_vrpsdptw}) & Co-GA \cite{WangC12} (with 12 parameters)\\
  \hline  
  \end{tabular}}
  \label{tab:summary}
\end{table*}

\section{Computational Studies}
\label{sec:experiment}

To assess the potential of CEPS, computational studies have been conducted with CEPS-TSP and CEPS-VRPSPDTW\footnote{The source code of CEPS-TSP and CEPS-VRPSPDTW, as well as the benchmark instances generated for the experiments, have been made available at \url{https://github.com/senshineL/CEPS}}.
The experiments mainly aim to address two questions:
\begin{enumerate}
  \item whether CEPS could better tackle the few-shots challenge, i.e., build generalizable PAPs with limited training instances, compared with the state-of-the-art PAP construction methods;
  \item whether co-evolution, i.e., alternately evolving the configuration population and the instance population, is effective as expected at enhancing the generalization of the resultant PAPs.
\end{enumerate}

To answer these two questions, two instance sets were firstly generated for TSP and VRPSPDTW, respectively.
The TSP instance set consists of 500 instances and the VRPSPDTW instance set consists of 233 instances.
It should be noted that, these instances are generated as our testbed.
To avoid bias towards CEPS, these instances should not be generated in the same way that CEPS evolves the instance population.
After the benchmark sets were generated, each of them was then randomly split into a training and a testing set, the size of which is 6\% and 94\% of the whole set, respectively.
To reduce the effect of the random splitting on the experimental results, the split was repeated for 3 times, leading to 3 unique pairs of training and testing sets for TSP and VRPSPDTW, denoted as TSP\_1/2/3 and VRPSPDTW\_1/2/3, respectively.
Throughout the experiments, testing instances were only used to approximate the generalization performance of the PAPs obtained by CEPS and compared methods.
Only the training instances were used for PAP construction, regardless of the methods used.
The TSP/VRPSPDTW instance set, the compared methods, and experimental protocol are further elaborated below. 

\subsection{Benchmark Instances}
For TSP, we collected 10 different instance generators from the literature, 
namely \textit{portgen}, \textit{ClusteredNetwork}, \textit{explosion}, \textit{implosion}, \textit{cluster}, \textit{rotation}, \textit{linearprojection}, \textit{expansion}, \textit{compression} and \textit{gridmutation}.
Among them \textit{portgen} generates a TSP instance (called \textit{rue} instance) by uniform randomly placing the points on a Euclidean plane.
It has been used to create test beds for the 8th DIMACS Implementation Challenge \cite{johnson2007experimental}.
The generator \textit{ClusteredNetwork} is from the R-package \textit{netgen} \cite{bossek2015netgen}, which generates an instance by placing points around different central points.
The other eight generators are proposed by a recent study \cite{BossekKN00T19}, which generate a TSP instance mainly by simulating a phenomenon in the point clouds of a \textit{rue} instance.
The details of these generators could be found in Appendix~\ref{appendix:generators}.
Considering the rather different instance-generation mechanisms underlying them, they are expected to generate highly-diverse TSP instances.
We used each of them to generate 50 instances, which finally gave us a set of 500 TSP instances.
The problem sizes (i.e., city number) of all these instances are 800.

For VRPSPDTW, we obtained data from a real-world application of the JD logistics company.
Specifically, the data contain customer requests that occurred during a period of time in a city.
The total number of customers is 3000, of which 400 customers require service per day.
Therefore, to generate a VRPSPDTW instance, we randomly select 400 customers from the 3000 customers, and the pickup/delivery demands of each customer are randomly selected from all the demands that the customer has during this period of time.
We repeated this process for 500 times, thus obtaining a set of 500 VRPSPDTW instances.
After that, a VRPSPDTW solver \cite{WangC12} was used to determine whether the generated instances have feasible solutions and those without feasible solutions were discarded.
Finally, we obtained a set of 233 VRPSPDTW instances.

\subsection{Compared Methods}
We compared CEPS with the state-of-the-art PAP construction methods (see Section~\ref{sec:related_work}), namely GLOBAL \cite{lindauer2017automatic}, PARHYDRA \cite{xu2010hydra,lindauer2017automatic} and PCIT \cite{LiuT019}.
It should be noted that all these methods involve no instance generation mechanism, i.e., the given training instances are assumed to sufficiently represent the target problem class.
Hence, given our experimental settings, comparison between CEPS and these approaches aims to evaluate whether CEPS could better tackle the few-shots challenge.

To address research question 2) raised at the beginning of Section~\ref{sec:experiment}, i.e., the role of co-evolution for achieving better (if any) generalization, a baseline method, named Evolution of Parameterized Search (EPS), was also adopted in the comparison.
EPS differs from CEPS in that it conducts instance mutation and PAP construction in two isolated phases, rather than alternately.
Given the same training instance set as CEPS, EPS first applies the instance mutation operator to generate an augmented set of instances.
The size of this augmented set is kept the same as the number of instances generated during the whole procedure of CEPS.
Then, a PAP is evolved with the same approach as in CEPS, but using the union of the initial and the augmented training instance sets as the input.
Moreover, to further verify the effectiveness of the co-evolution in CEPS, we included the initial PAPs of CEPS (the PAPs built by the initialization phase, lines 2-7 in Algorithm~\ref{alg:ceps}) in the comparison with the final PAPs obtained by CEPS.

\subsection{Experimental Protocol}
We set the number of component solvers in PAP, i.e., $K$, to 4, since 4-core machines are widely available now.
The parameters of the compared methods were set following suggestions in the literature.
For CEPS, the number of iterations of the co-evolution, i.e., \textit{MaxIte}, and the number of temporary PAPs generated, i.e., $n$, were set to 4 and 10, respectively.
The termination condition for the evolution of the instance population in CEPS was the predefined time budget being exhausted. 
In the experiments the total CPU time consumed by each compared method was kept almost the same.
The detailed time settings of each compared method are presented in Appendix~\ref{appendix:timesetting}.
The above-described experimental settings, as well as the used parameterized algorithms and performance indicators (see Section~\ref{sec:instantiation}), are summarized in Table~\ref{tab:summary}.

For each pair of training and testing sets, i.e., TSP\_1/2/3 and VRPSPDTW\_1/2/3, we applied each considered method to construct a PAP on the training set and then tested the resulting PAP on the testing set.
For each testing instance, the PAP was applied for 3 runs and the median of those three runs was recorded as the performance of the solver on the instance.
The performance of a PAP on different testing instances were then aggregated to obtain the number of total timeouts (\#TOs), PAR-10 (for TSP solver) and PANC (for VRPSPDTW solver) on the testing set.
All the experiments were conducted on a cluster of 3 Intel Xeon machines with 128 GB RAM and 24 cores each (2.20 GHz, 30 MB Cache), running Centos 7.5.

\subsection{Results and Analysis}

\begin{table*}
\centering
\caption{The testing results of the PAPs constructed by each method. \#TOs refers to number of the total timeouts. PAR-10 and PANC are penalized average runtime-10 and penalized average normalized cost, respectively.
Performance of a PAP is highlighted in grey if it achieved the best testing performance.}
\scalebox{1.15}{
\begin{tabular}{l|cc|cc|cc|cc|cc|cc}
    \hline
    \multirow{2}[2]{*}{} & \multicolumn{2}{c|}{TSP-1} & \multicolumn{2}{c|}{TSP-2} & \multicolumn{2}{c|}{TSP-3} & \multicolumn{2}{c|}{VRPSPDTW-1} & \multicolumn{2}{c|}{VRPSPDTW-2} & \multicolumn{2}{c}{VRPSPDTW-3} \\
          & \#TOs & PAR-10 & \#TOs & PAR-10 & \#TOs & PAR-10 & \#TOs & PANC  & \#TOs & PANC  & \#TOs & PANC \\
    \hline
    GLOBAL & 10    & 3.85  & 15    & 5.18  & 10    & 3.67  & 4     & 253   & 3     & 248   & 4     & 258 \\
    PCIT  & \cellcolor{gray!50}6     & \cellcolor{gray!50}2.51 & \cellcolor{gray!50}4     & 2.44  & 9     & 3.46  & 4     & 258   & 2     & 240   & 6     & 274 \\
    PARHYDRA & 9     & 3.55  & \cellcolor{gray!50}4     & 2.19  & 5     & 2.36  & 2     & 237   & 5     & 265   & 3     & 249 \\
    EPS   & 7    & 2.93   & 6    & 2.81  & 8     & 2.81  & 2     & 238   & 2     & 236   & \cellcolor{gray!50}1     & \cellcolor{gray!50}229 \\
    CEPS.initial & 14 & 4.50 & 14 & 4.63 & 6 & 3.12 & 3 & 245 & 2 & 237 & 2 & 237 \\
    CEPS & \cellcolor{gray!50}6     & 2.74 & \cellcolor{gray!50}4     & \cellcolor{gray!50}2.15 & \cellcolor{gray!50}2     & \cellcolor{gray!50}1.94 & \cellcolor{gray!50}0     & \cellcolor{gray!50}221 & \cellcolor{gray!50}1     & \cellcolor{gray!50}229 & \cellcolor{gray!50}1     & \cellcolor{gray!50}229 \\
    \hline
\end{tabular}}
\label{tab:main_results}
\end{table*}

\begin{figure*}[t]
  \centering
    \begin{subfigure}[b]{0.325\linewidth}
     \centering
     \scalebox{1.0}{\includegraphics[width=\linewidth]{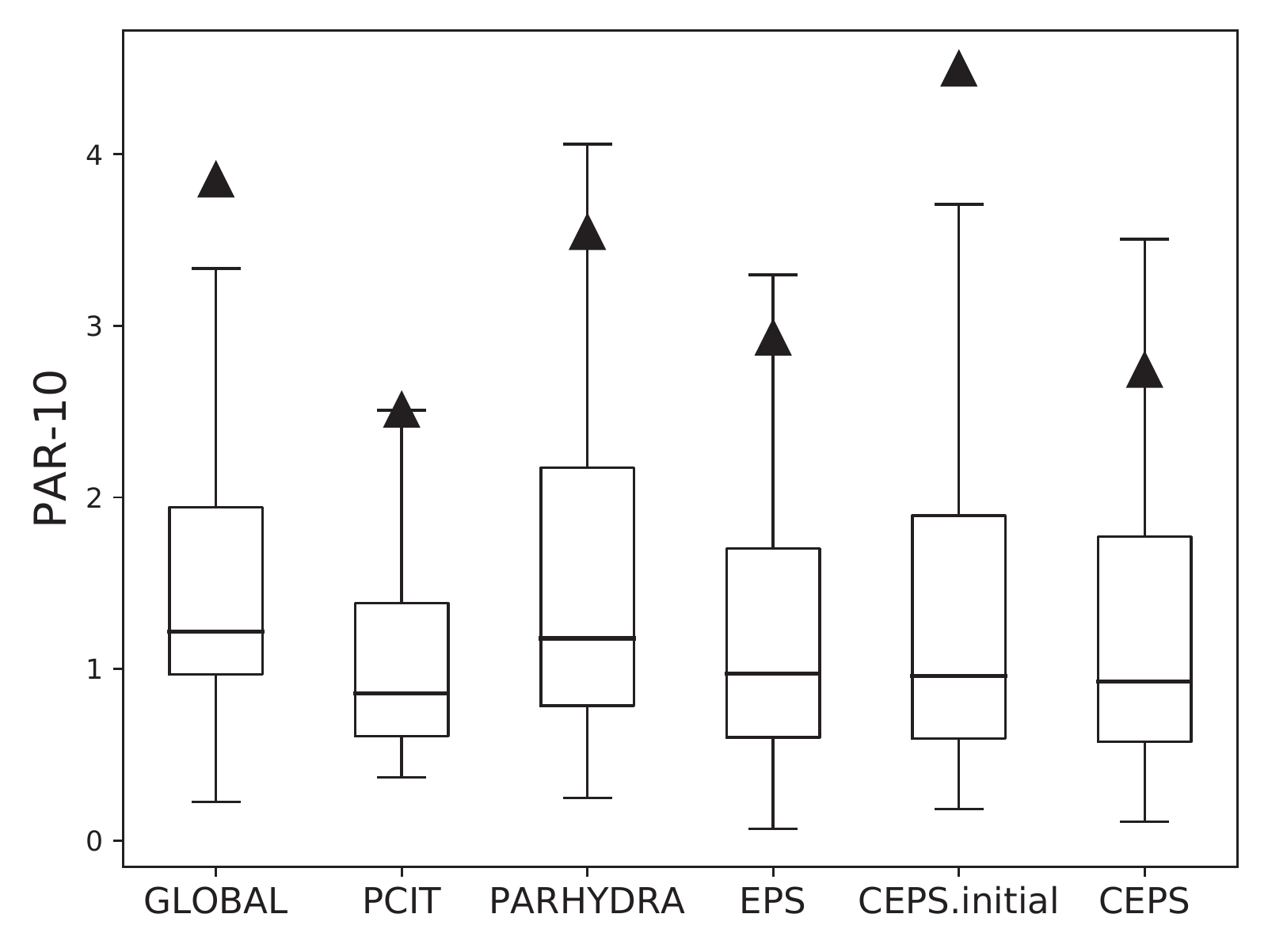}}
     \caption{TSP-1}
     \label{fig:boxplot_tsp_1}
    \end{subfigure}
    \hfill
    \begin{subfigure}[b]{0.325\linewidth}
     \centering
      \scalebox{1.0}{\includegraphics[width=\linewidth]{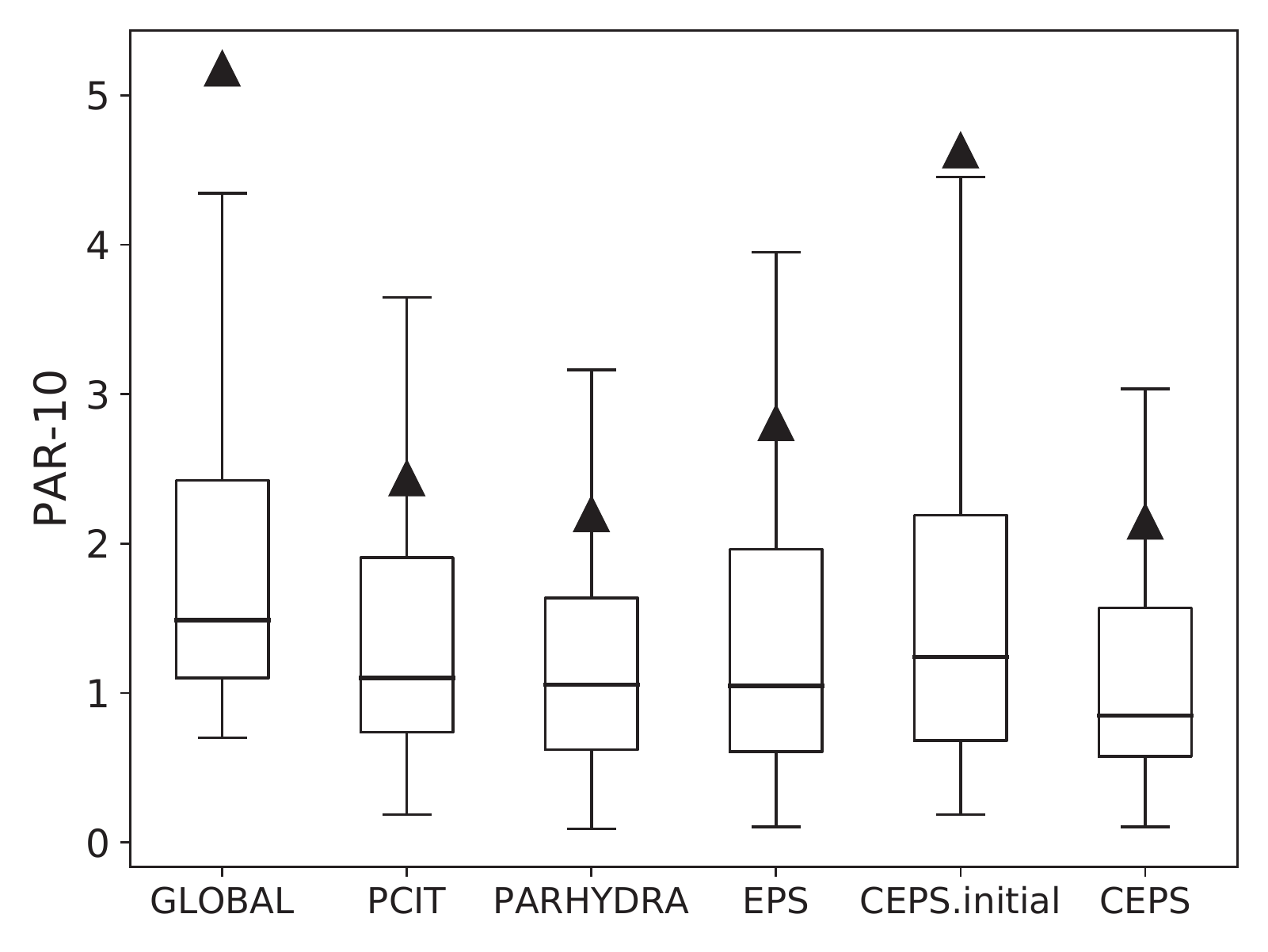}}
      \caption{TSP-2}
      \label{fig:boxplot_tsp_2}
    \end{subfigure}
    \hfill
    \begin{subfigure}[b]{0.325\linewidth}
     \centering
      \scalebox{1.0}{\includegraphics[width=\linewidth]{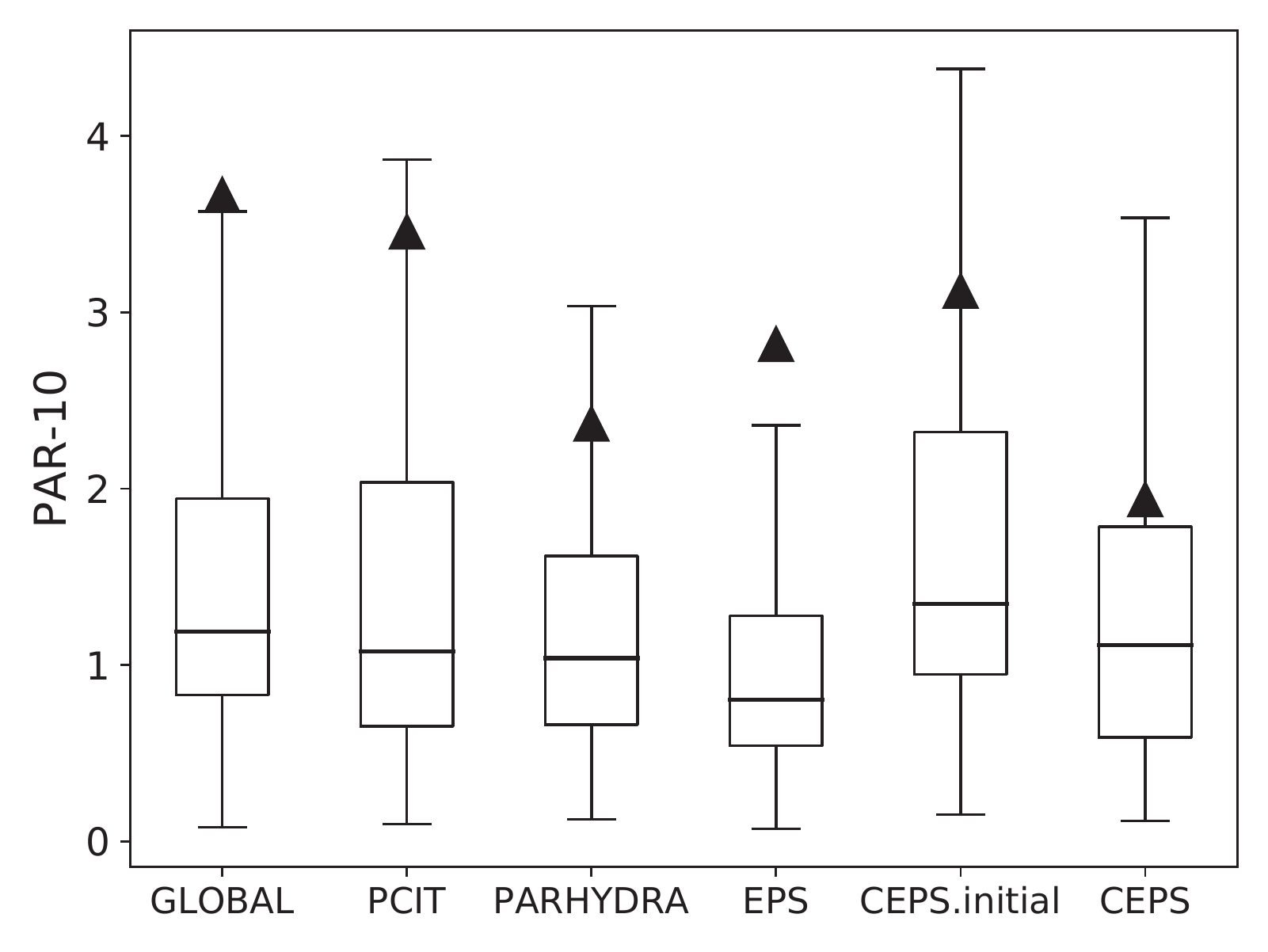}}
      \caption{TSP-3}
      \label{fig:boxplot_tsp_3}
    \end{subfigure}

    \begin{subfigure}[b]{0.325\linewidth}
      \centering
      \scalebox{1.0}{\includegraphics[width=\linewidth]{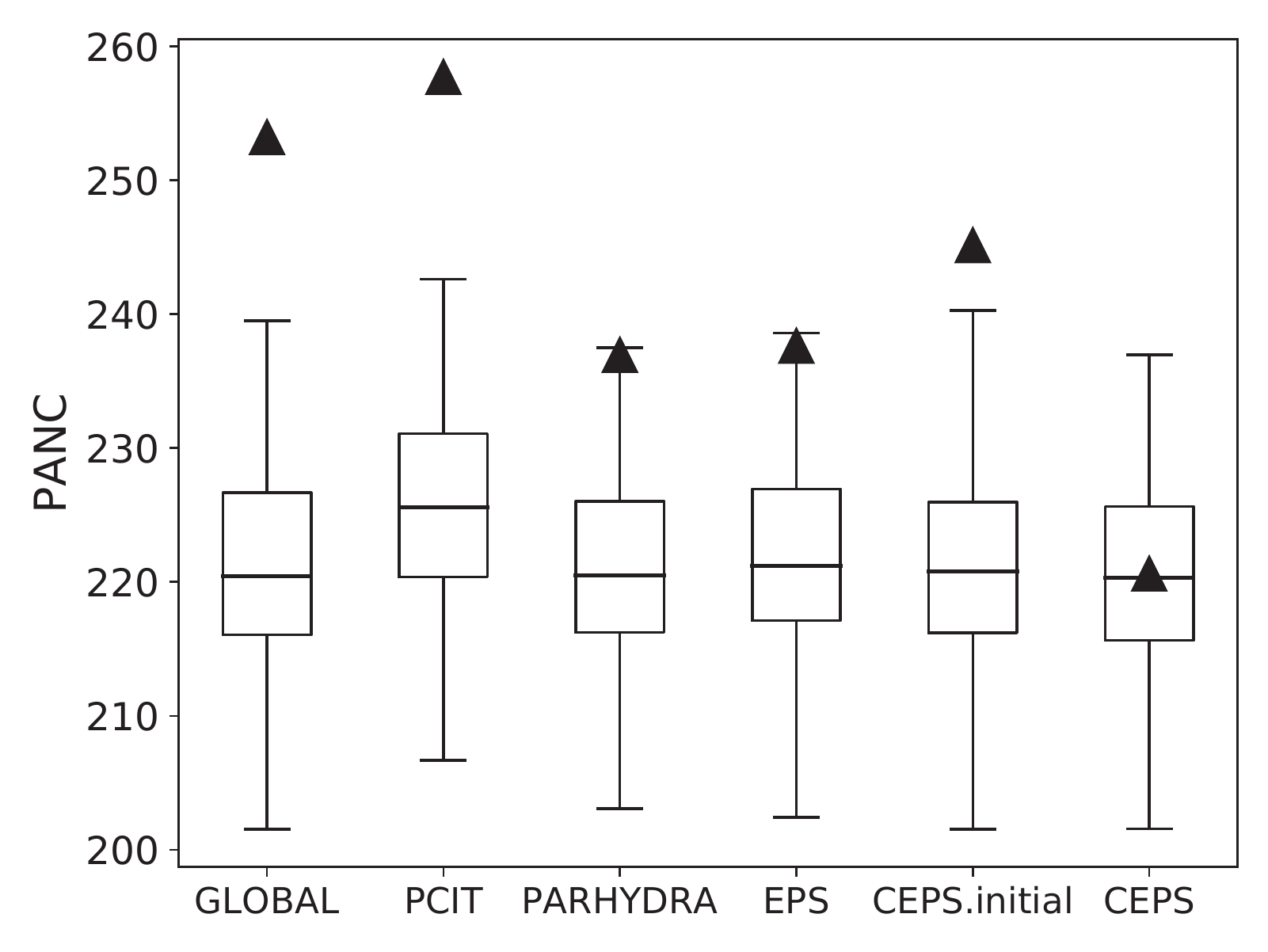}}
      \caption{VRPSPDTW-1}
      \label{fig:boxplot_vrpspdtw_1}
     \end{subfigure}
     \hfill
     \begin{subfigure}[b]{0.325\linewidth}
      \centering
       \scalebox{1.0}{\includegraphics[width=\linewidth]{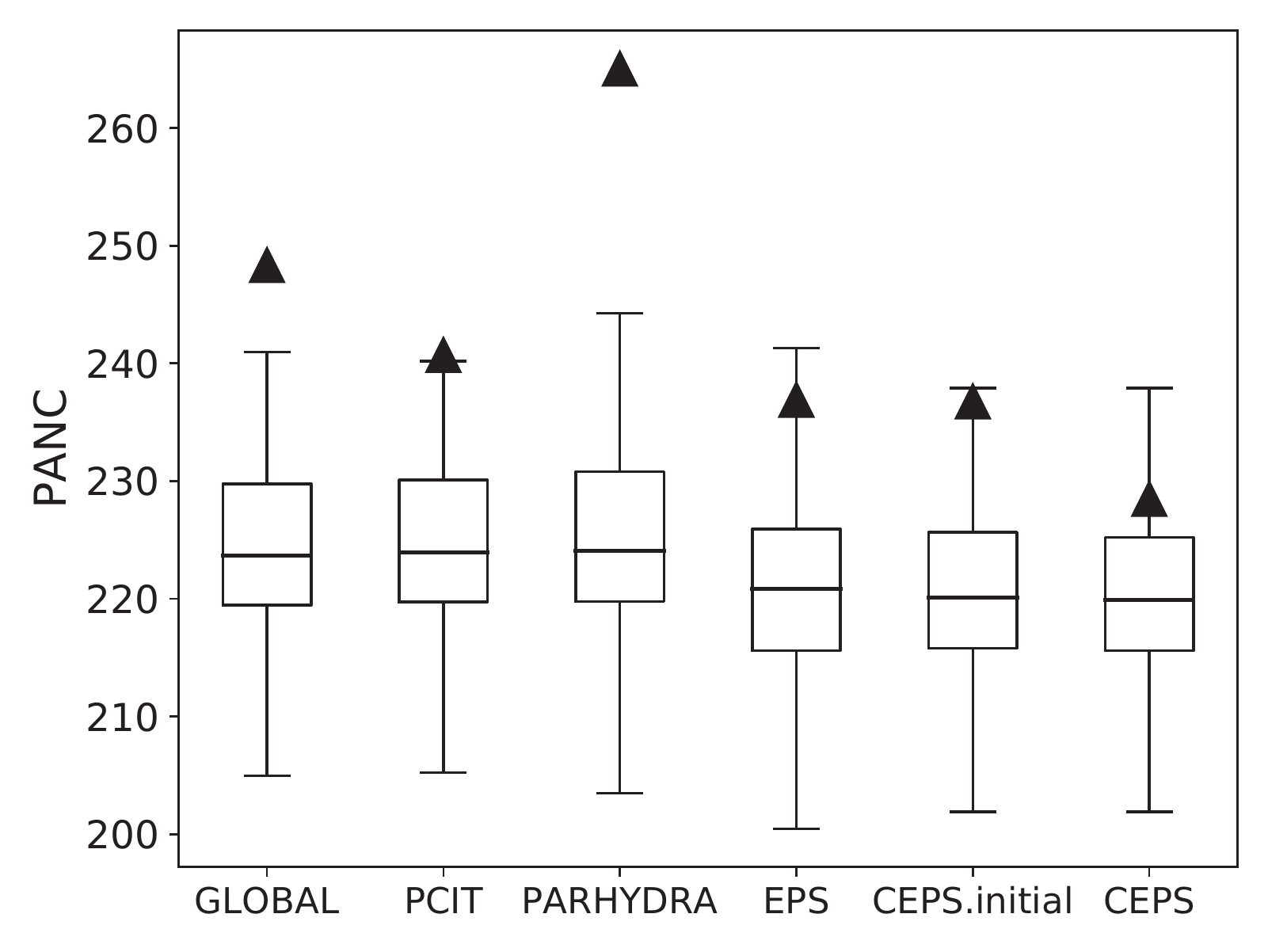}}
       \caption{VRPSPDTW-2}
       \label{fig:boxplot_vrpspdtw_2}
     \end{subfigure}
     \hfill
     \begin{subfigure}[b]{0.325\linewidth}
      \centering
       \scalebox{1.0}{\includegraphics[width=\linewidth]{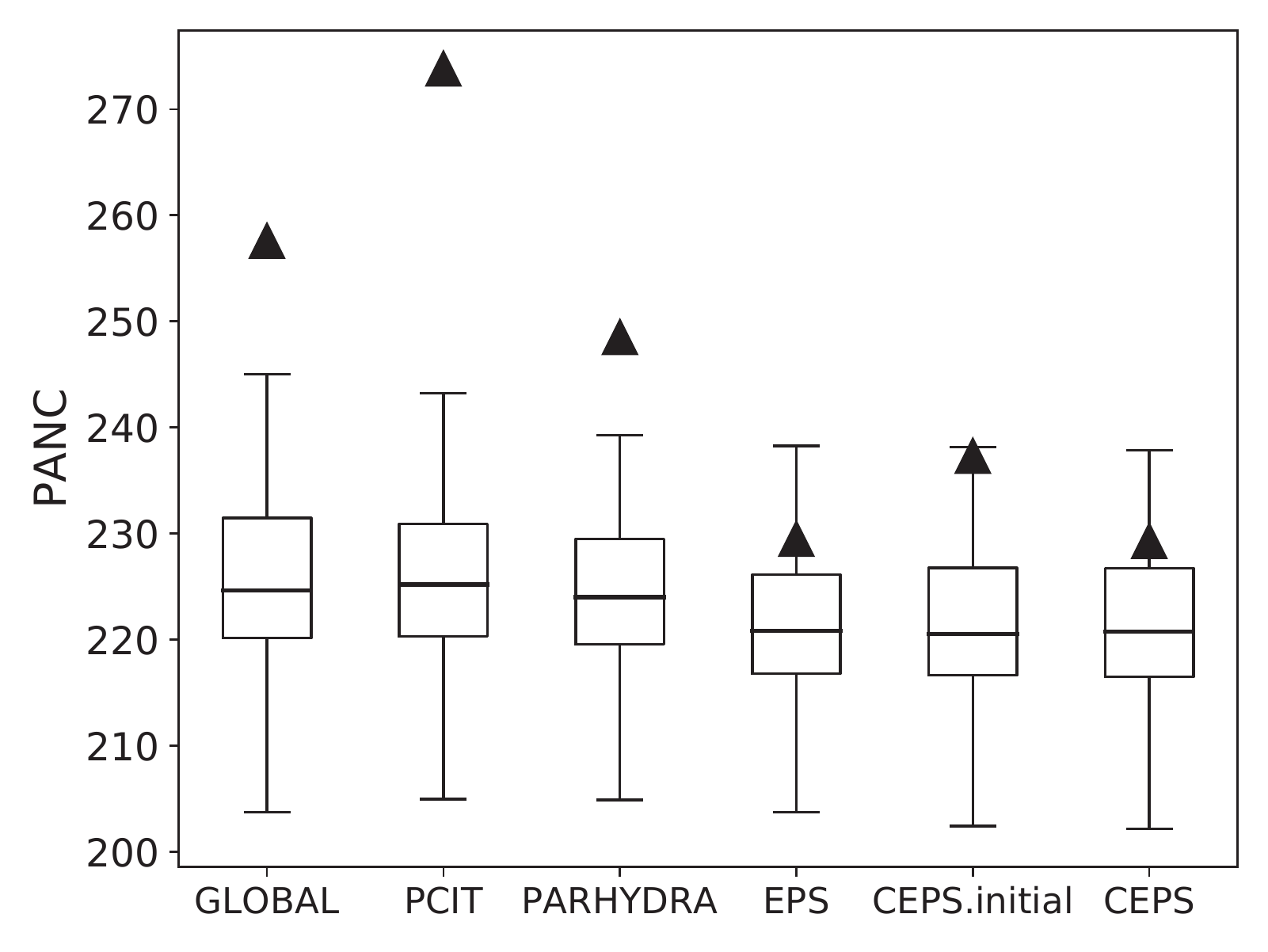}}
       \caption{VRPSPDTW-3}
       \label{fig:boxplot_vrpspdtw_3}
     \end{subfigure}

  \caption{Visual comparison in boxplots of the medians and variance of the test performance of each PAP across the testing instances. Note the mean value is also plotted, indicated by ``$\blacktriangle$''.}
  \label{fig:boxplots}
\end{figure*}

We report the \#TOs, PAR-10 and PANC achieved by the PAPs on the testing set in Table~\ref{tab:main_results} and
also visualize their medians and variance across all the testing instances by boxplots in Figure~\ref{fig:boxplots}.
Note the mean value is also plotted in Figure \ref{fig:boxplots} (indicated by ``$\blacktriangle$'') to show that for a PAP how its PAR-10/PANC is affected by the outliers (the timeout cases) which would be hidden by boxplots.
In Table~\ref{tab:main_results} the \#TOs, PAR-10/PANC of a PAP is highlighted in grey if it achieved the best performance.
%and is indicated in bold face if it was not significantly different from the best performance on the corresponding testing set.
% Note for PAR-10/PANC, the performance is better if the score is smaller.
One could make three important observations from these results. 
First, the PAPs obtained by CEPS have the smallest number of timeouts in all the six experiments, which means they have the highest success rate for solving the testing instances among all the tested PAPs.
Recall that CEPS actively searches in the instance space to identify the hard-to-solve instances for further improving the generalization of the PAPs.
Such a mechanism makes CEPS the method that is least affected by the hard testing instances which significantly differs from the given training instances.
This could be further verified by Figure~\ref{fig:boxplots}, in which CEPS is the method that has the least gap between the mean value (which takes timeouts into account) and median value (which naturally filters out the timeouts).
Moreover, thanks to the least number of timeouts, in five out of the six experiments, the PAPs output by CEPS achieved the best scores in terms of PAR-10 and PANC.
Typically, one could observe that in TSP-1 and TSP-3 of Figure~\ref{fig:boxplots}, PARHDYRA and EPS have ``better'' performances than CEPS on the normal instances, but finally achieved worse PAR-10 due to more timeouts.
Furthermore, recall that in different experiments the training/testing splits were different, compared to other approaches, CEPS performed more stably over all 6 experiments.
For instance, the \#TOs of PCIT and PARHYDRA fluctuates over different training/testing sets on VRPSPDTW problem.
In summary, CEPS is not only the best-performing method, but also is less sensitive to the training data, i.e., could better tackle the few-shots challenge.

Second, EPS also involves instance generation, while was outperformed by methods that do not generate synthetic instances in several cases, e.g., compared to PARHYDRA on TSP\_2.
This observation indicates that isolating instance generation from PAP construction may have negative effects.
On the other hand, the fact CEPS performed better than EPS shows the effectiveness of integrating instance generation into the co-evolving framework.

Third, compared to the initial PAPs of CEPS (indicated by ``CEPS.initial'' in Table~\ref{tab:main_results}), the final PAPs obtained by CEPS performed better on all of the six experiments.
On average, the performance improvement rate (in terms of PAR-10 and PANC) is 21.78\%.
These results indicate that the co-evolution in CEPS is effective as expected at enhancing the generalization of the PAP solvers.

\subsection{Comparison with State-of-the-art TSP solvers}

\begin{figure*}[t]
  \centering
    \begin{subfigure}[b]{0.325\linewidth}
     \centering
     \scalebox{1.0}{\includegraphics[width=\linewidth]{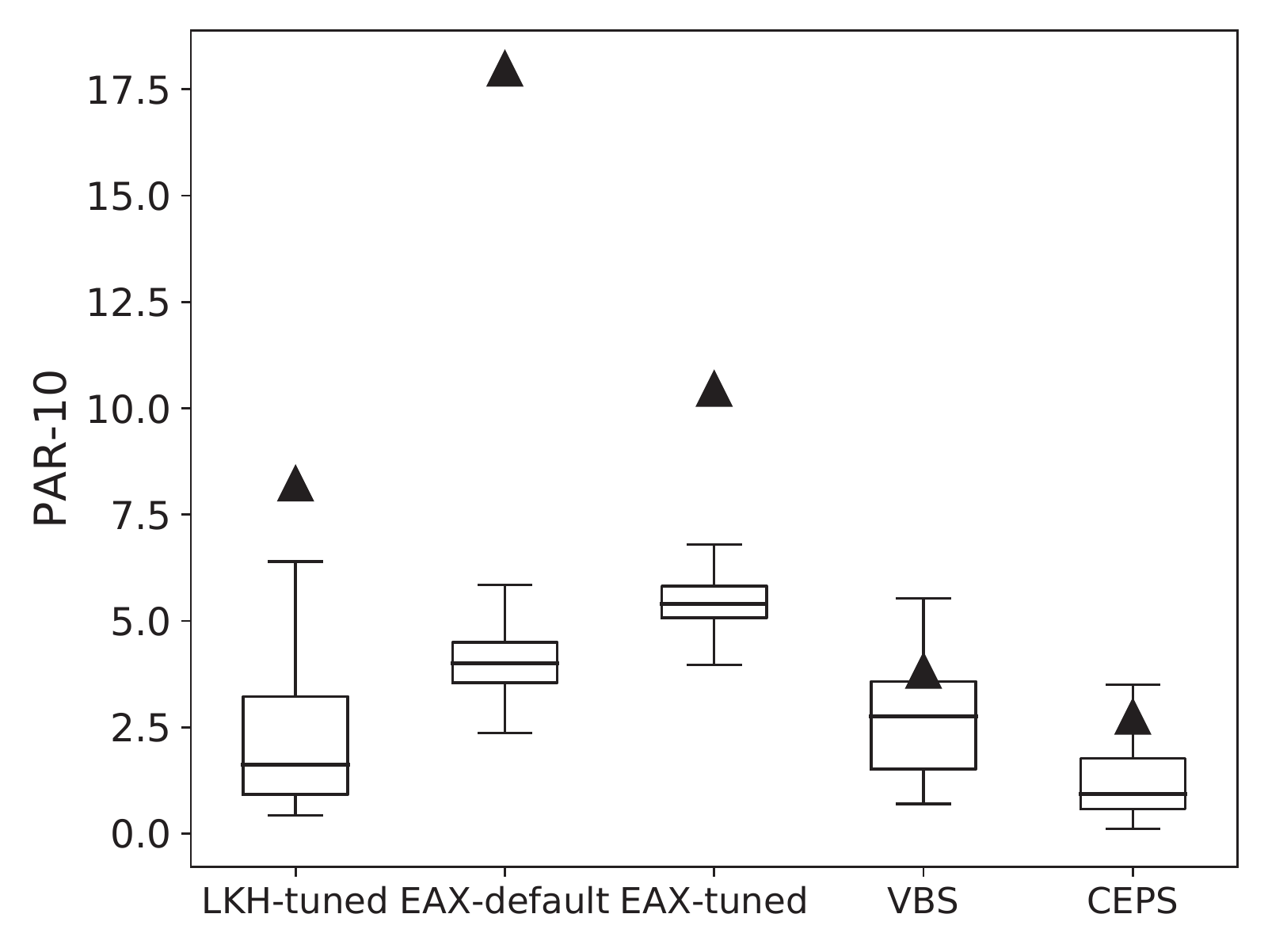}}
     \caption{TSP-1}
     \label{fig:boxplot_tsp_sota_1}
    \end{subfigure}
    \hfill
    \begin{subfigure}[b]{0.325\linewidth}
     \centering
      \scalebox{1.0}{\includegraphics[width=\linewidth]{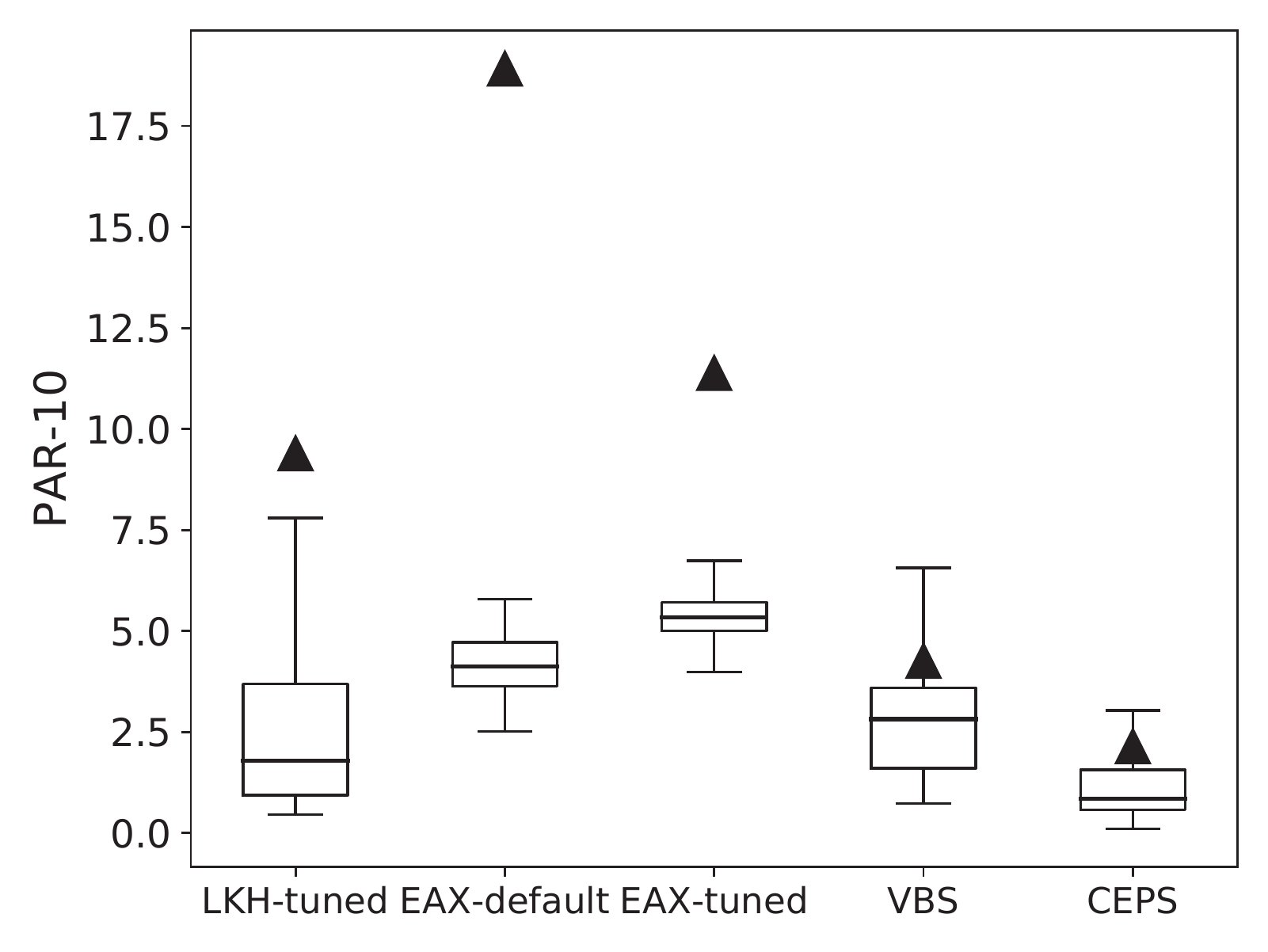}}
      \caption{TSP-2}
      \label{fig:boxplot_tsp_sota_2}
    \end{subfigure}
    \hfill
    \begin{subfigure}[b]{0.325\linewidth}
     \centering
      \scalebox{1.0}{\includegraphics[width=\linewidth]{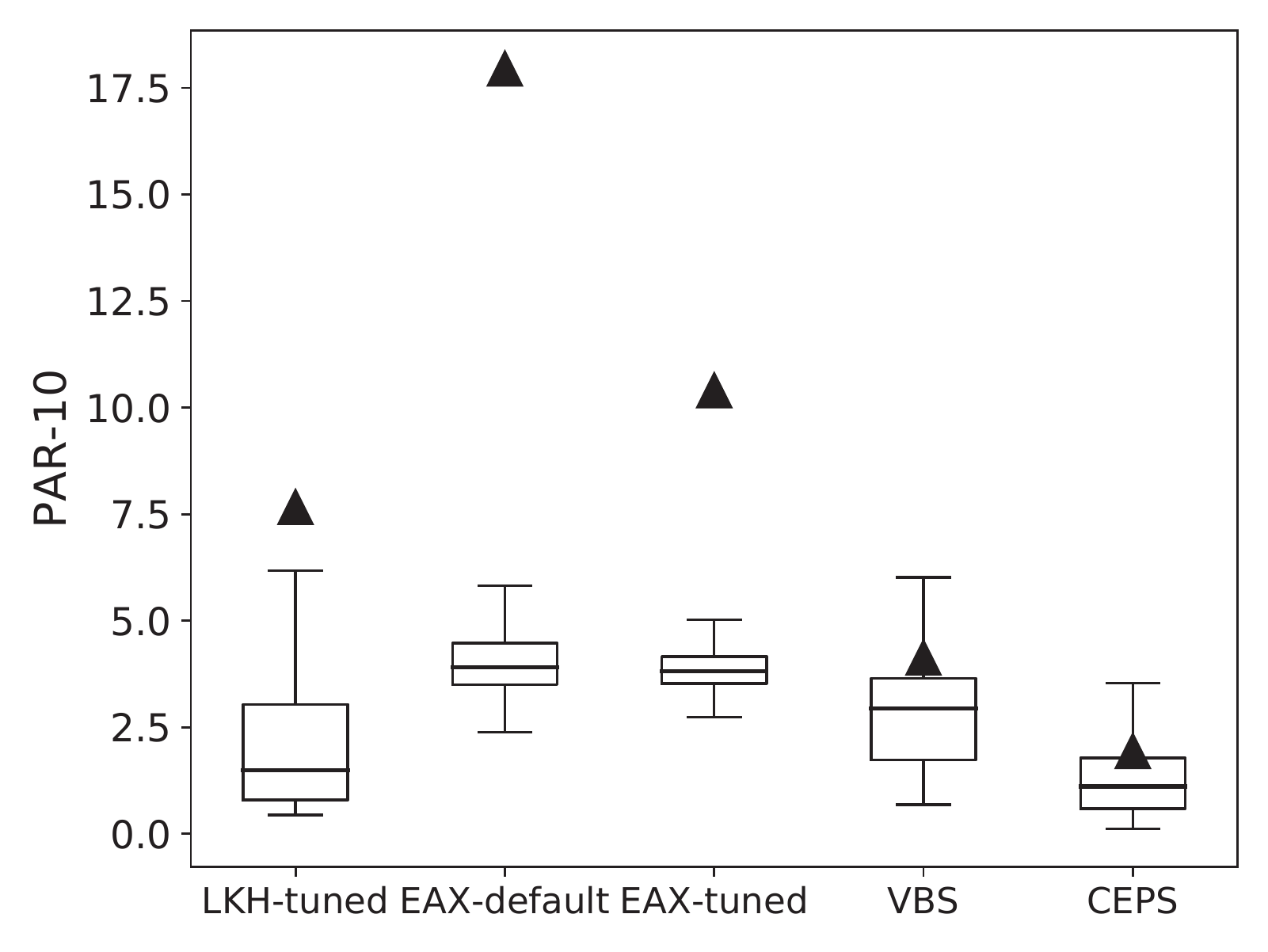}}
      \caption{TSP-3}
      \label{fig:boxplot_tsp_sota_3}
    \end{subfigure}
  \caption{Visual comparison in boxplots of the medians and variance of the test performance of each TSP solver across the testing instances. Note the mean value is also plotted, indicated by ``$\blacktriangle$''.}
  \label{fig:boxplot_tsp_sota}
\end{figure*}

For CEPS-TSP, to further assess its performance, we compared the PAPs constructed by it with the state-of-the-art TSP solvers.
More specifically, we considered:
1) the default configuration of the considered LKH algorithm, denoted as LKH-default;
2) the default configuration of another powerful TSP algorithm, EAX \cite{nagata2013powerful}, denoted as EAX-default, which has been demonstrated to outperform LKH on a broad range of TSP instances;
%3) the default configuration of a variant of EAX, namely EAX-restart(default) \cite{Dubois-LacosteH15}, which is original EAX augmented with a restart mechanism to handle the premature termination issue;
3) the tuned versions of LKH and EAX, denoted as LKH-tuned and EAX-tuned, respectively, which are obtained by running SMAC on their configuration spaces and the training sets for the same time budget of the PAP construction of CEPS.
In addition, we also considered two state-of-the-art portfolio-based algorithm selection (AS) methods for TSP \cite{KerschkeKBHT18,zhao2020leveraging} (see Section~\ref{sec:related_work}).
Since these two methods have adopted the same TSP algorithm portfolio which contains LKH, EAX, MAOS \cite{XieL09} and their variants, instead of comparing each of these two method with CEPS, we directly adopted the virtual best solver (VBS) of their algorithm portfolio.
VBS is the \textit{oracle} or \textit{perfect} selector which always chooses the best algorithm for each instance without any selecting cost.
This idealized procedure provides an upper bound for the performance of any algorithm selector;
due to imperfect selection and the cost incurred by selecting, VBS cannot be achieved in practice by actual algorithm selectors.

As before, for each testing instance, we applied each solver for 3 runs and the median of those three runs was recorded as the performance of the solver on the instance.
We report the \#TOs and PAR-10 achieved by these solvers on the testing set in Table~\ref{tab:tsp_sota} and also visualize their medians and variance across all the testing instances by boxplots in Figure~\ref{fig:boxplot_tsp_sota}.
Note LKH-default is omitted in Figure~\ref{fig:boxplot_tsp_sota} due to its large number of timeouts.
There are two important findings from these results.
First, LKH-default and EAX-default performed badly on the testing set, with considerable timeouts.
We speculate that this is because the default configurations of LKH and EAX are mainly designed to handle much larger-scale TSP instances (e.g., 10000) than the instances considered here.
After being tuned (i.e., LKH-tuned and EAX-tuned), they both achieved significant performance improvement, though still obviously falling behind of the PAPs obtained by CEPS.
Second, the only solver that could match the PAP's performance level in one of the three scenarios, is the VBS of the algorithm portfolio considered by the algorithm selection approaches \cite{KerschkeKBHT18,zhao2020leveraging}.
However, in TSP-2 and TSP-3, the performance advantage of the PAP is still significant.
%Note the algorithm portfolio for these approaches is built manually by human experts.
%Besides, VBS and PAP are equivalent in the sense that their performance on any given instance is the best performance achieved among its component solvers on the instance.
%Thus the above results that PAP outperformed VBS of the portfolio considered in \cite{KerschkeKBHT18,zhao2020leveraging} basically means the algorithm portfolio obtained by CEPS is better than the one built by human experts, which further indicates the great potential of building algorithm portfolios via automatic construction approaches such as CEPS, instead of building manually.

\begin{table}[t]
 \centering
 \caption{Comparison of the state-of-the-art TSP solvers with the PAPs obtained by CEPS, on the testing set. \#TOs refers to number of total timeouts. PAR-10 is the penalized average runtime-10. Performance of a solver is highlighted in grey if it achieved the best testing performance.}
 \scalebox{0.95}{
 \begin{tabular}{l|cc|cc|cc}
     \hline
     \multirow{2}[2]{*}{} & \multicolumn{2}{c|}{TSP-1} & \multicolumn{2}{c|}{TSP-2} & \multicolumn{2}{c}{TSP-3} \\
           & \#TOs & PAR-10 & \#TOs & PAR-10 & \#TOs & PAR-10 \\
     \hline
     LKH-default & 131    & 30.84  & 137    & 31.98  & 150   & 34.73 \\
     LKH-tuned  & 29    & 8.23  & 34    & 9.40  & 27   & 7.67 \\
     EAX-default   & 69    & 17.98   & 73    & 18.91  & 69     & 17.95 \\
     EAX-tuned & 33 & 10.97 & 30 & 10.38 & 29 & 10.12 \\
     VBS & \cellcolor{gray!50}6     & 3.82 & 7     & 4.26 & 6   & 4.13 \\
     CEPS & \cellcolor{gray!50}6     & \cellcolor{gray!50}2.74 & \cellcolor{gray!50}4     & \cellcolor{gray!50}2.15 & \cellcolor{gray!50}2     & \cellcolor{gray!50}1.94 \\
     \hline
 \end{tabular}}
 \label{tab:tsp_sota}
\end{table}

\subsection{Assessing Generalization on Existing VRPSPDTW Benchmarks}
\begin{table}[tbph]
  \centering
  \caption{Comparison between the solutions found by PAP and the best-known solutions (BKS) found before (as reported in the literature) on existing VRPSPDTW benchmark instances. \#better, \#not-worse and \#worse refers to the number of the instances on which PAP found better, not worse (i.e., either with better or the same quality) and worse solutions compared to the BKS.}
    \begin{tabular}{c|c|c|c|c}
    \hline
    Instance Type &  \#instances & \#better & \#not-worse & \#worse \\
    \hline
    RCdp (small) & 9     & 0 (0\%) & 9 (100\%) & 0 (0\%) \\
    Rdp   & 23    & 4 (17\%) & 17 (57\%) & 6 (26\%) \\
    Cdp   & 17    & 0 (0\%) & 9 (53\%) & 8 (47\%) \\
    RCdp  & 16    & 6 (13\%) & 8 (38\%) & 8 (50\%) \\
    \hline
    count  &  65   & 10 (15\%) & 43 (66 \%) & 22 (34\%)\\
    \hline
    \end{tabular}%
\label{tab:results_vrpspdtw}%
\end{table}%

To further investigate the generalization ability of CEPS, 
the PAP constructed by CEPS in the VRPSPDTW\_1 experiment have been tested on existing VRPSPDTW benchmarks \cite{WangC12}, which is a widely used benchmark for VRPSPDTW.
Note that compared to the benchmarks, the training instances in VRPSPDTW\_1 were obtained from different sources (i.e., real-world application), and may have quite different problem characteristics, e.g., customer number and node distribution.
Hence, the PAP constructed by CEPS could be said to generalize well to totally unseen data if it was constructed using the VRPSPDTW\_1 training set while still performing well on the VRPSPDTW benchmarks.
Table~\ref{tab:results_vrpspdtw} presents the comparisons between the solutions found by the PAP and the best-known solutions reported in the literature \cite{WangC12,WangMZS15,wulanhuang2016,XingchengP2018,Yongshi2018} (up to May 2019), regardless of what algorithm was used. 
Table 3 shows that overall the PAP could generalize well to the existing benchmarks.
On 43 out of 65 (66\%) instances, the solutions found by the PAP are not worse than the best solutions currently known.
It is notable that on 10 instances the PAP found new best solutions.
Another observation is that the PAP performed not very well on the ``cdp'' type instances, in which the locations of customers are clustered (see \cite{WangC12} for details).
We speculate that this is because the parameterized algorithm used for tuning PAPs has an inherent deficiency when handling this type of instances, which on the other hand indicates that highly parameterized algorithms with flexible solving capacities are important to fully exploit the power of CEPS on a specific problem class. 

\section{Threats to Validity}
\label{sec:threats}
There are several validity threats to the findings of this study.
The first one is the correctness of the implementation of all the compared methods.
Prior to commencing our experiments, we have thoroughly checked the source code of these methods (obtained from online or implemented by ourselves) and ensured that the implementations were correct.

Second, the results of our experiments are limited to the data sets used, in which the TSP instances are generated by ten different generators while the VRPSPDTW instances are collected from real-world application.
We have made these instances available online.
In the future we will assess CEPS-TSP and CEPS-VRPSPDTW on more instance sets obtained from other sources (generators and applications).

Third, although we have demonstrated that CEPS could better tackle the few-shots challenge than existing PAP construction methods in two case studies, there is no guarantee that CEPS could be easily applied to other problem domains.
Actually, an instantiation of CEPS to a specific domain involves specification of instance mutation operator, the parameterized algorithm, as well as the automatic algorithm configuration method.
Hence, as demonstrated by CEPS-TSP and CEPS-VRPSPDTW, one needs to first define these three modules according to previous literature on the target problem class, or from scratch.
Among these three modules, the instance generators could be adapted from one problem to another more easily and the automatic algorithm configuration methods are usually generic.
Hence, the parameterized search method might be the most crucial (also the most difficult-to-obtain) one among the three modules.

% Although CEPS is not restricted to a specific type of search methods, the PAP-type method adopted in this work is suggested as a default choice, not only for its advantages in terms of solution quality (compared to a single solver) on a large variety of problems \cite{xu2010hydra,PengTCY10,lindauer2017automatic,LiuT019,LiuT020} and the theoretical guarantee it offers as elaborated in Section~\ref{sec:theoretical_insights}, but also because PAP allows exploiting modern high-performance computing facilities in an extremely simple way.
% This merit is sometimes even more important than solution quality, since the wall-clock runtime is always a crucial performance indicator for real-world optimization systems.

\section{Conclusion}
\label{sec:conclusion}
In this work, a co-evolutionary approach, i.e., CEPS, is proposed for constructing  PAPs to obtain good generalization performance.
By co-evolving the training instance set and the configurations, CEPS gradually guides the search of configurations towards instances on which the current configurations fail to perform well, and thus leads to PAPs that could generalize better.
From a theoretical point of view, the evolution of instance set is essentially a greedy mechanism for instance augmentation that guarantees the generalization performance of the resultant solver to improve as much as possible.
As a result, CEPS is particularly effective in case that only a limited number of problem instances is available.
Such a scenario is usually true when building real-world systems for tackling hard optimization problems.
Two concrete instantiations, i.e., CEPS-TSP and CEPS-VRPSPDTW, are also presented.
The performance of the two instantiations on TSP and VRPSPDTW problems support the effectiveness of CEPS in the sense that, in comparison with state-of-the-art PAP construction approaches, the PAPs obtained by CEPS achieves better generalization performance.

Since CEPS is a generic framework, some discussions would help elaborate issues that are of significance in practice.
First, although this work assumes CEPS takes a set of initial training instances as the input, such training instances are not necessarily real-world instances but could be generated randomly.
In other words, CEPS could be used in a fully cold-start setting (a.k.a. zero-shot), i.e., no real-world instances are available for the target problem class.
Further, CEPS could either be run offline or online, i.e., it could accommodate new real instances whenever available.

Second, the potential of CEPS could be further explored by taking advantage of the data generated during its run, except for the final obtained PAP.
The data contain all the sampled configurations and instances, and the performance of the former on the latter.
Considering that when using a search method to solve a problem instance, its optimal parameter values are usually problem-instance dependent and thus need to be tuned.
To tune parameters for a new problem instance, we can learn from the historical data generated by CEPS to build a mapping from problem instances to their optimal parameter values, i.e., a low-cost online parameter-tuning system for any single instance.
It could be seen as an extension of the common algorithm selection systems \cite{Kotthoff14} which select the best algorithm/configuration for a  given instance from a predefined algorithm set.
In addition, the challenging instance sets generated by CEPS could be further used on comprehensive analysis of the strengths and weaknesses of the parameterized algorithms, such as for which configurations are those instances challenging and further improvement of the algorithm.
We have made the instance sets generated by CEPS available online to further facilitate the investigations on them.

Finally, it is interesting to note that the emerging topic of learn to optimize, which explores utilizing machine learning techniques, e.g., reinforcement learning, to build neural networks for solving optimization problems \cite{NazariOST18,ChenT19,KoolHW19,MnihKSRVBGRFOPB15,KulkarniNST16}, could also be combined with CEPS.
In this case, the implementation of CEPS would be able to leverage on gradient descent methods to tune/evolve the configurations (i.e., training the weights of a network).

% or


\appendices

\section{TSP Instance Generators}
\label{appendix:generators}
The adopted 10 TSP generators include the \textit{portgen} generator from the 8th DIMACS Implementation Challenge \cite{johnson2007experimental}, the \textit{ClusteredNetwork} generator from the R-package \textit{netgen} \cite{bossek2015netgen} and 8 TSP instance generators, namely \textit{explosion}, \textit{implosion}, \textit{cluster}, \textit{rotation}, \textit{linearprojection}, \textit{expansion}, \textit{compression} and \textit{gridmutation}, from the R-package \textit{tspgen} \cite{BossekKN00T19}.

\begin{enumerate}
\item The \textit{portgen} generator generates an instance by uniform randomly placing the points. The generated instances are called \textit{rue} instances.
\item The \textit{ClusteredNetwork} generator generates an instance by placing points around different central points. The number of the clusters were set to 4,5,6,7, and 8, for each of which 10 instances were generated.
\item The \textit{explosion} generator generates an instance by tearing holes into the city points of a \textit{rue} instance, with all points within the explosion range pushed out of the explosion area.
\item The \textit{implosion} generator generates an instance by driving the city points of a \textit{rue} instance towards a randomly sampled implosion center.
\item The \textit{cluster} generator generates an instance by randomly sampling a cluster centroid in a \textit{rue} instance, and then moving a randomly selected set of points into the cluster region.
\item The \textit{rotation} generator generates an instance by rotating a subset of points of a \textit{rue} instance with a randomly selected angle.
\item The \textit{linearprojection} generator generates an instance by projecting a subset of points of a \textit{rue} instance to a linear function.
\item The \textit{expansion} generator generates an instance by placing a tube around a linear function in the points of a \textit{rue} instance, and then orthogonally pushes all points within that tube out of that region.
\item The \textit{compression} generator generates an instance by squeezing a set of randomly selected points of a \textit{rue} instance from within a tube (surrounding a linear function) towards the tube’s central axis.
\item The \textit{gridmutation} generator generates an instance by randomly relocating a ``box'' of city points of a \textit{rue} instance.
\end{enumerate}

\section{Detailed Time Settings of Compared Methods}
\label{appendix:timesetting}
\begin{table}[t]
  \centering
  \caption{Detailed time settings (in hours) of each PAP construction method.}
  \begin{tabular}{lllllc}
  \hline
  & \multicolumn{5}{c}{TSP} \\
  \hline
  & $t_{c}$ & $t_{v}$ & $t_{i}$ & $t_{ini}$ & CPU Time \\
  \hline
  CEPS & 1.5h & 0.5h  & 1.5h & 8h &  320h\\
  GLOBAL & 7.5h & 1h & -- & -- & 340h \\
  PCIT   & 7.5h & 1h & -- & -- & 340h \\
  PARHYDRA &2h & 1h  & -- & -- & 300h \\
  \hline
  & \multicolumn{5}{c}{VRPSPDTW} \\
  \hline
  & $t_{c}$ & $t_{v}$ & $t_{i}$ & $t_{ini}$ & CPU Time \\
  \hline
  CEPS &  6h  & 2h     & 6h & 32h & 1312h\\
  GLOBAL &   30h     & 4h     & -- & --  & 1360h\\
  PCIT   &   30h     & 4h     & -- & --  & 1360h\\
  PARHYDRA & 8h  & 4h      & --  & -- & 1200h\\
  \hline
\end{tabular}
\label{tab:timesetting}%
\end{table}

The most time-consuming parts of PAP construction methods are the runs of the configurations on the problem instances, and the incurred computational costs account for the vast majority of the total costs.
For CEPS, the configurations would be run in the initialization phase (line 5 in Algorithm~\ref{alg:ceps}), in the evolution of the configuration population (line 12 and line 15 in Algorithm~\ref{alg:ceps}) and in the evolution of the instance population (line 22 in Algorithm~\ref{alg:ceps}).
Therefore for each of these four procedures we set the corresponding wall-clock time budget, i.e., $t_{init}$, $t_c$, $t_v$ and $t_i$, to control the overall computational costs of CEPS.
Then the total CPU time consumed by CEPS could be estimated by $t_{init} + MaxIte \cdot K \cdot\left[n \cdot\left(t_{c}+t_{v}\right) + t_i\right]$.
In this paper, $K$, $MaxIte$ and $n$ are set to 4, 4 and 10, respectively.

The total CPU time consumed by GLOBAL and PCIT could be estimated by 
$K \cdot n \cdot \left(t_c + t_v \right)$, while for PARHYDRA it could be estimated by $\Sigma_{i=1}^K i \cdot n \cdot \left(t_c + t_v\right) $(see \cite{LiuT019,xu2010hydra,lindauer2017automatic} for how these results are derived).
Note for different methods $t_c$, $t_v$ and $t_i$ could be set to different values.
The detailed setting of the time budget for each PAP construction method is given in Table~\ref{tab:timesetting}.
Overall the total CPU time consumed by each method is kept almost the same.



% \section*{Acknowledgment}
% This work is supported in part by the Guangdong Provincial Key Laboratory (Grant No. 2020B121201001), the Program for Guangdong Introducing Innovative and Entrepreneurial Teams (Grant No. 2017ZT07X386), the Shenzhen Peacock Plan (Grant No. KQTD2016112514355531), the Science and Technology Commission of Shanghai Municipality (No. 19511120600), the National Leading Youth Talent Support Program of China, and the MOE University Scientific-Technological Innovation Plan Program.
\ifCLASSOPTIONcaptionsoff
  \newpage
\fi



%\bibliographystyle{IEEEtran}


% \end{thebibliography}

\bibliographystyle{IEEEtran}
\bibliography{IEEEabrv,mybib}

\begin{thebibliography}{10}
\providecommand{\url}[1]{#1}
\csname url@samestyle\endcsname
\providecommand{\newblock}{\relax}
\providecommand{\bibinfo}[2]{#2}
\providecommand{\BIBentrySTDinterwordspacing}{\spaceskip=0pt\relax}
\providecommand{\BIBentryALTinterwordstretchfactor}{4}
\providecommand{\BIBentryALTinterwordspacing}{\spaceskip=\fontdimen2\font plus
\BIBentryALTinterwordstretchfactor\fontdimen3\font minus
  \fontdimen4\font\relax}
\providecommand{\BIBforeignlanguage}[2]{{%
\expandafter\ifx\csname l@#1\endcsname\relax
\typeout{** WARNING: IEEEtran.bst: No hyphenation pattern has been}%
\typeout{** loaded for the language `#1'. Using the pattern for}%
\typeout{** the default language instead.}%
\else
\language=\csname l@#1\endcsname
\fi
#2}}
\providecommand{\BIBdecl}{\relax}
\BIBdecl

\bibitem{LinK73}
S.~Lin and B.~W. Kernighan, ``{An effective heuristic algorithm for the
  traveling-salesman Problem},'' \emph{Operations Ressarch}, vol.~21, no.~2,
  pp. 498--516, 1973.

\bibitem{EibenS11}
A.~E. Eiben and S.~K. Smit, ``{Parameter tuning for configuring and analyzing
  evolutionary algorithms},'' \emph{Swarm and Evolutionary Computation},
  vol.~1, no.~1, pp. 19--31, 2011.

\bibitem{hutter2011sequential}
F.~Hutter, H.~H. Hoos, and K.~Leyton{-}Brown, ``{Sequential model-based
  optimization for general algorithm configuration},'' in \emph{Proceedings of
  the 5th International Conference on Learning and Intelligent Optimization,
  {LION}'2011}.\hskip 1em plus 0.5em minus 0.4em\relax Rome, Italy: Springer,
  Jan 2011, pp. 507--523.

\bibitem{KarafotiasHE15}
G.~Karafotias, M.~Hoogendoorn, and {\'{A}}.~E. Eiben, ``{Parameter control in
  evolutionary algorithms: trends and challenges},'' \emph{IEEE Transactions on
  Evolutionary Computation}, vol.~19, no.~2, pp. 167--187, 2015.

\bibitem{HuangLY20}
C.~Huang, Y.~Li, and X.~Yao, ``{A survey of automatic parameter tuning methods
  for metaheuristics},'' \emph{IEEE Transactions on Evolutionary Computation},
  vol.~24, no.~2, pp. 201--216, 2020.

\bibitem{DymondEKH15}
A.~S.~D. Dymond, A.~P. Engelbrecht, S.~Kok, and P.~S. Heyns, ``Tuning
  optimization algorithms under multiple objective function evaluation
  budgets,'' \emph{IEEE Transactions on Evolutionary Computation}, vol.~19,
  no.~3, pp. 341--358, 2015.

\bibitem{hutter2009paramils}
F.~Hutter, H.~H. Hoos, K.~Leyton{-}Brown, and T.~St{\"{u}}tzle, ``{ParamILS: An
  automatic algorithm configuration framework},'' \emph{Journal of Artificial
  Intelligence Research}, vol.~36, no.~1, pp. 267--306, 2009.

\bibitem{ansotegui2009gender}
C.~Ans{\'{o}}tegui, M.~Sellmann, and K.~Tierney, ``{A gender-based genetic
  algorithm for the automatic configuration of algorithms},'' in
  \emph{Proceedings of the 15th International Conference on Principles and
  Practice of Constraint Programming, {CP}'2009}.\hskip 1em plus 0.5em minus
  0.4em\relax Lisbon, Portugal: Springer, Sep 2009, pp. 142--157.

\bibitem{AnsoteguiMSST15}
C.~Ans{\'{o}}tegui, Y.~Malitsky, H.~Samulowitz, M.~Sellmann, and K.~Tierney,
  ``{Model-based genetic algorithms for algorithm configuration},'' in
  \emph{Proceedings of the 24th International Joint Conference on Artificial
  Intelligence, {IJCAI}'2015}.\hskip 1em plus 0.5em minus 0.4em\relax Buenos
  Aires, Argentina: {AAAI} Press, Jul 2015, pp. 733--739.

\bibitem{lopez2016irace}
M.~L{\'o}pez-Ib{\'a}{\~n}ez, J.~Dubois-Lacoste, L.~{P{\'e}rez C{\'a}ceres},
  T.~St{\"u}tzle, and M.~Birattari, ``{The irace package: Iterated racing for
  automatic algorithm configuration},'' \emph{Operations Research
  Perspectives}, vol.~3, pp. 43--58, 2016.

\bibitem{GomesS01}
C.~P. Gomes and B.~Selman, ``{Algorithm portfolios},'' \emph{Artificial
  Intelligence}, vol. 126, no. 1-2, pp. 43--62, 2001.

\bibitem{huberman1997economics}
B.~A. Huberman, R.~M. Lukose, and T.~Hogg, ``{An economics approach to hard
  computational problems},'' \emph{Science}, vol. 275, no. 5296, pp. 51--54,
  1997.

\bibitem{LiuT019}
S.~Liu, K.~Tang, and X.~Yao, ``{Automatic construction of parallel portfolios
  via explicit instance grouping},'' in \emph{Proceedings of the 33rd {AAAI}
  Conference on Artificial Intelligence, {AAAI}' 2019}.\hskip 1em plus 0.5em
  minus 0.4em\relax Honolulu, HI: {AAAI} Press, Jan 2019, pp. 1560--1567.

\bibitem{PengTCY10}
F.~Peng, K.~Tang, G.~Chen, and X.~Yao, ``{Population-based algorithm portfolios
  for numerical optimization},'' \emph{{IEEE} Transactions on Evolutionary
  Computation}, vol.~14, no.~5, pp. 782--800, 2010.

\bibitem{LiuT020}
S.~Liu, K.~Tang, and X.~Yao, ``{Generative adversarial construction of parallel
  portfolios},'' \textit{IEEE Transactions on Cybernetics}, 2020, to be
  published, DOI:10.1109/TCYB.2020.2984546.

\bibitem{asanovic2006landscape}
K.~Asanovic, R.~Bod{\'{\i}}k, J.~Demmel, T.~Keaveny, K.~Keutzer,
  J.~Kubiatowicz, N.~Morgan, D.~A. Patterson, K.~Sen, J.~Wawrzynek, D.~Wessel,
  and K.~A. Yelick, ``{A view of the parallel computing landscape},''
  \emph{Communications of the ACM}, vol.~52, no.~10, pp. 56--67, 2009.

\bibitem{Reinelt91}
G.~Reinelt, ``{TSPLIB} - {A} traveling salesman problem library,''
  \emph{INFORMS Journal on Computing}, vol.~3, no.~4, pp. 376--384, 1991.

\bibitem{WangC12}
H.~Wang and Y.~Chen, ``{A genetic algorithm for the simultaneous delivery and
  pickup problems with time window},'' \emph{Computers \& Industrial
  Engineering}, vol.~62, no.~1, pp. 84--95, 2012.

\bibitem{NazariOST18}
M.~Nazari, A.~Oroojlooy, L.~V. Snyder, and M.~Tak{\'{a}}c, ``{Reinforcement
  learning for solving the vehicle routing problem},'' in \emph{Proceedings of
  the 31st Annual Conference on Neural Information Processing Systems,
  {NeurIPS}'2018}.\hskip 1em plus 0.5em minus 0.4em\relax Quebec, Canada:
  Curran Associates Inc., Dec 2018, pp. 9861--9871.

\bibitem{ChenT19}
X.~Chen and Y.~Tian, ``{Learning to perform local rewriting for combinatorial
  optimization},'' in \emph{Proceedings of the 32ed Annual Conference on Neural
  Information Processing Systems, {NeurIPS}'2019}.\hskip 1em plus 0.5em minus
  0.4em\relax Vancouver, Canada: Curran Associates Inc., Dec 2019, pp.
  6278--6289.

\bibitem{KoolHW19}
W.~Kool, H.~van Hoof, and M.~Welling, ``{Attention, Learn to solve routing
  problems!}'' in \emph{Proceedings of the 7th International Conference on
  Learning Representations, {ICLR}'2019}.\hskip 1em plus 0.5em minus
  0.4em\relax New Orleans, LA: OpenReview.net, May 2019.

\bibitem{BlotHJKT16}
A.~Blot, H.~H. Hoos, L.~Jourdan, M.~Kessaci{-}Marmion, and H.~Trautmann,
  ``{MO-ParamILS: {A} multi-objective automatic algorithm configuration
  framework},'' in \emph{Proceedings of the 10th International Conference on
  Learning and Intelligent Optimization, {LION}'2016}.\hskip 1em plus 0.5em
  minus 0.4em\relax Ischia, Italy: Springer, Jun 2016, pp. 32--47.

\bibitem{Birattari2004}
M.~Birattari, ``{On the estimation of the expected performance of a
  metaheuristic on a class of instances},'' Technical Report TR/IRIDIA/2004-01,
  IRIDIA, Universit{\'e} Libre de Bruxelles, Brussels, Belgium, Tech. Rep.,
  2004.

\bibitem{LiuTL020}
S.~Liu, K.~Tang, Y.~Lei, and X.~Yao, ``On performance estimation in automatic
  algorithm configuration,'' in \emph{Proceedings of the 34th AAAI Conference
  on Artificial Intelligence, {AAAI}' 2020}.\hskip 1em plus 0.5em minus
  0.4em\relax New York, NY: {AAAI} Press, Feb 2020, pp. 2384--2391.

\bibitem{lindauer2017automatic}
M.~Lindauer, H.~H. Hoos, K.~Leyton{-}Brown, and T.~Schaub, ``{Automatic
  construction of parallel portfolios via algorithm configuration},''
  \emph{Artificial Intelligence}, vol. 244, pp. 272--290, 2017.

\bibitem{xu2010hydra}
L.~Xu, H.~Hoos, and K.~Leyton{-}Brown, ``{Hydra: Automatically configuring
  algorithms for portfolio-based selection},'' in \emph{Proceedings of the 24th
  {AAAI} Conference on Artificial Intelligence, {AAAI'}2010}.\hskip 1em plus
  0.5em minus 0.4em\relax Atlanta, GA: {AAAI} Press, Jul 2010, pp. 210--216.

\bibitem{kadioglu2010isac}
S.~Kadioglu, Y.~Malitsky, M.~Sellmann, and K.~Tierney, ``{ISAC -
  Instance-specific algorithm configuration},'' in \emph{Proceedings of the
  19th European Conference on Artificial Intelligence, {ECAI}'2010}.\hskip 1em
  plus 0.5em minus 0.4em\relax Lisbon, Portugal: {IOS} Press, Aug 2010, pp.
  751--756.

\bibitem{xu2008satzilla}
L.~Xu, F.~Hutter, H.~H. Hoos, and K.~Leyton{-}Brown, ``{SATzilla:
  Portfolio-based algorithm selection for SAT},'' \emph{Journal of Artificial
  Intelligence Research}, vol.~32, pp. 565--606, 2008.

\bibitem{KerschkeKBHT18}
P.~Kerschke, L.~Kotthoff, J.~Bossek, H.~H. Hoos, and H.~Trautmann,
  ``{Leveraging TSP solver complementarity through machine learning},''
  \emph{Evolutionary Computation}, vol.~26, no.~4, pp. 597--620, 2018.

\bibitem{zhao2020leveraging}
K.~Zhao, S.~Liu, Y.~Rong, and J.~X. Yu, ``{Leveraging TSP solver
  complementarity via deep learning},'' \emph{arXiv preprint arXiv:2006.00715},
  2020.

\bibitem{Kotthoff14}
L.~Kotthoff, ``{Algorithm selection for combinatorial search problems: {A}
  survey},'' \emph{{AI} Magazine}, vol.~35, no.~3, pp. 48--60, 2014.

\bibitem{RosinB97}
C.~D. Rosin and R.~K. Belew, ``New methods for competitive coevolution,''
  \emph{Evolutionary Computation}, vol.~5, no.~1, pp. 1--29, 1997.

\bibitem{Hemert06}
J.~I. van Hemert, ``{Evolving combinatorial problem instances that are
  difficult to solve},'' \emph{Evolutionary Computation}, vol.~14, no.~4, pp.
  433--462, 2006.

\bibitem{Helsgaun09}
K.~Helsgaun, ``{General \emph{k}-opt submoves for the Lin-Kernighan {TSP}
  heuristic},'' \emph{Mathematical Programming Computation}, vol.~1, no. 2-3,
  pp. 119--163, 2009.

\bibitem{johnson2007experimental}
D.~S. Johnson and L.~A. McGeoch, ``{Experimental analysis of heuristics for the
  {STSP}},'' in \emph{{The Traveling Salesman Problem and Its Variations}},
  G.~Gutin and A.~P. Punnen, Eds.\hskip 1em plus 0.5em minus 0.4em\relax
  Springer, 2007, pp. 369--443.

\bibitem{bossek2015netgen}
J.~Bossek, ``{netgen: Network generator for combinatorial graph problems},''
  https://github.com/jakobbossek/netgen, 2015.

\bibitem{BossekKN00T19}
J.~Bossek, P.~Kerschke, A.~Neumann, M.~Wagner, F.~Neumann, and H.~Trautmann,
  ``{Evolving diverse TSP instances by means of novel and creative mutation
  operators},'' in \emph{Proceedings of the 15th {ACM/SIGEVO} Conference on
  Foundations of Genetic Algorithms, {FOGA}'2019}.\hskip 1em plus 0.5em minus
  0.4em\relax Potsdam, Germany: {ACM}, Aug 2019, pp. 58--71.

\bibitem{nagata2013powerful}
Y.~Nagata and S.~Kobayashi, ``{A powerful genetic algorithm using edge assembly
  crossover for the traveling salesman problem},'' \emph{INFORMS Journal on
  Computing}, vol.~25, no.~2, pp. 346--363, 2013.

\bibitem{XieL09}
X.~Xie and J.~Liu, ``{Multiagent optimization system for solving the traveling
  salesman problem (tsp)},'' \emph{{IEEE} Transactions on Systems, Man, and
  Cybernetics, Part {B}}, vol.~39, no.~2, pp. 489--502, 2009.

\bibitem{WangMZS15}
C.~Wang, D.~Mu, F.~Zhao, and J.~W. Sutherland, ``{A parallel simulated
  annealing method for the vehicle routing problem with simultaneous
  pickup-delivery and time windows},'' \emph{Computers \& Industrial
  Engineering}, vol.~83, pp. 111--122, 2015.

\bibitem{wulanhuang2016}
W.~Huang and T.~Zhang, ``{Vehicle routing problem with simultaneous pick-up and
  delivery and time-windows based on improved global artificial fish swarm
  algorithm},'' \emph{Computer Engineering and Applications}, vol.~52, no.~21,
  pp. 21--29, 2016.

\bibitem{XingchengP2018}
X.~Pu and K.~Wang, ``{An evolutionary ant colony algorithm for a vehicle
  routing problem with simultaneous pick-up and delivery and hard time
  windows},'' in \emph{Proceedings of the 30th Chinese Control and Decision
  Conference, {CCDC}'2018}.\hskip 1em plus 0.5em minus 0.4em\relax Shenyang,
  China: IEEE, Jun 2018, pp. 6499--6503.

\bibitem{Yongshi2018}
Y.~Shi, T.~Boudouh, and O.~Grunder, ``{An efficient tabu search based procedure
  for simultaneous delivery and pick-up problem with time window},''
  \emph{IFAC-PapersOnLine}, vol.~51, no.~11, pp. 241--246, 2018.

\bibitem{MnihKSRVBGRFOPB15}
V.~Mnih, K.~Kavukcuoglu, D.~Silver, A.~A. Rusu, J.~Veness, M.~G. Bellemare,
  A.~Graves, M.~A. Riedmiller, A.~Fidjeland, G.~Ostrovski, S.~Petersen,
  C.~Beattie, A.~Sadik, I.~Antonoglou, H.~King, D.~Kumaran, D.~Wierstra,
  S.~Legg, and D.~Hassabis, ``{Human-level control through deep reinforcement
  learning},'' \emph{Nature}, vol. 518, no. 7540, pp. 529--533, 2015.

\bibitem{KulkarniNST16}
T.~D. Kulkarni, K.~Narasimhan, A.~Saeedi, and J.~Tenenbaum, ``{Hierarchical
  deep reinforcement learning: integrating temporal abstraction and intrinsic
  motivation},'' in \emph{Proceedings of the 29th Annual Conference on Neural
  Information Processing Systems, {NIPS}'2016}.\hskip 1em plus 0.5em minus
  0.4em\relax Barcelona, Spain: Curran Associates Inc., Dec 2016, pp.
  3675--3683.

\end{thebibliography}


%\vfill


\end{document}